\documentclass{article}

\usepackage[preprint,nonatbib]{neurips_2020}

\usepackage[utf8]{inputenc} 
\usepackage[T1]{fontenc}    
\usepackage{url}            
\usepackage{booktabs}       
\usepackage{amsfonts}       
\usepackage{nicefrac}       
\usepackage{microtype}      
\usepackage{epsfig}
\usepackage{graphicx}
\usepackage{amsmath}
\usepackage{amssymb}
\usepackage{ascii}
\usepackage[T1]{fontenc}
\usepackage{mathrsfs}
\usepackage{booktabs}
\usepackage{multirow}
\usepackage{array}
\usepackage{colortbl}
\usepackage{color}
\usepackage{lipsum}
\usepackage{graphicx}
\usepackage{mathtools}
\usepackage{bm}
\usepackage{cite}
\usepackage{subfigure}
\usepackage{enumerate}
\usepackage{multicol}
\usepackage[titletoc]{appendix}
\usepackage[pagebackref=true,breaklinks=true,letterpaper=true,colorlinks,citecolor=blue,bookmarks=false]{hyperref}

\title{OccInpFlow: Occlusion-Inpainting Optical Flow Estimation by Unsupervised Learning}
\author{%
  Kunming Luo$^1$ \qquad
  Chuan Wang$^1$ \qquad
  Nianjin Ye$^{1,2}$ \qquad
  Shuaicheng Liu$^{1,2}$ \qquad
  Jue Wang$^1$ \\
  $^1$Megvii Technology \\
  $^2$University of Electronic Science and Technology of China
  \\
  \texttt{\{luokunming,wangchuan,yenianjin,liushuaicheng,wangjue\}@megvii.com}
}

\begin{document}

\maketitle

\begin{abstract}\label{sec:abs}
Occlusion is an inevitable and critical problem in unsupervised optical flow learning. Existing methods either treat occlusions equally as non-occluded regions or simply remove them to avoid incorrectness. However, the occlusion regions can provide effective information for optical flow learning. In this paper, we present OccInpFlow, an occlusion-inpainting framework to make full use of occlusion regions. Specifically, a new appearance-flow network is proposed to inpaint occluded flows based on the image content. Moreover, a boundary warp is proposed to deal with occlusions caused by displacement beyond image border. We conduct experiments on multiple leading flow benchmark data sets such as Flying Chairs, KITTI and MPI-Sintel, which demonstrate that the performance is significantly improved by our proposed occlusion handling framework. 
\end{abstract}
\section{Introduction}
Optical flow estimation is a fundamental vision task~\cite{Simonyan2014,bonneel2015blind,KITTI_2015}. There are various challenges regarding accurate flow estimation, such as large displacements, noisy interferences, motion details and occlusions. Traditional approaches optimize hand-crafted energy functions to obtain flow fields~\cite{Sun2010,Sun2014,Sun2016,Anurag2019,Kroeger2016,Hur2017MirrorFlowES}. Learning based methods, on the other hand, directly estimate flows from images by convolutional neural networks (CNNs)~\cite{Flownet_flyingchairs}. Various network designs have been proposed to overcome these problem-inherit challenges~\cite{FlowNet2,semiflow_2017_nips,unsup_CPN,spynet2017,LiteFlowNet} to improve the performance.

Among all these challenges, the most difficult one is the occlusion, which is the key issue that prevent high quality performance and practical applications. Occlusion comes from dynamic objects, discontinues depth disparities, and frame boundaries. Traditional methods often regularize incoherent motion to propagate flow from non-occluded surroundings to occluded parts~\cite{Berthold1981,Thomas2004,Thomas2015,EpicFlow_2015}. However, this could fail when occlusion regions are wrongly located or motions propagated from incorrect motion sources. Whereas, training supervised CNNs requires large amount of annotated training data~\cite{KITTI_2012,KITTI_2015,Flownet_flyingchairs,Butler2012}, which is hard to obtain in practice, especially when there are occlusions~\cite{Alvarez2007,Ayvaci2010,Sun2010nips,Sun2014cvpr}.

On the other hand, unsupervised optical flow can learn from unlabeled data by minimizing photometric loss between images, such that the training data is no longer a bottleneck. However, directly minimizing photometric loss cannot learn correct flows in occlusion regions due to incorrect warped results. Unfortunately, early works of unsupervised family simply ignored the influence of occlusions and treat them equally as non-occlusion regions~\cite{unsup_ICIP_2016,Epipolar_flow_2019cvpr,Ren2017aaai,Jason2016}. To handle this problem, Wang~\emph{et al.} proposed an occlusion aware learning where occlusions are detected and discarded during loss calculation~\cite{wang2018}. DDFlow~\cite{Pengpeng2019} and SelFlow~\cite{Liu2019CVPR} applied random cropping to hallucinate occlusions for self-supervision. However, all these methods miss the guidance in real occlusion regions.

\begin{figure}[t]
  \centering
  \includegraphics[width=0.98\linewidth]{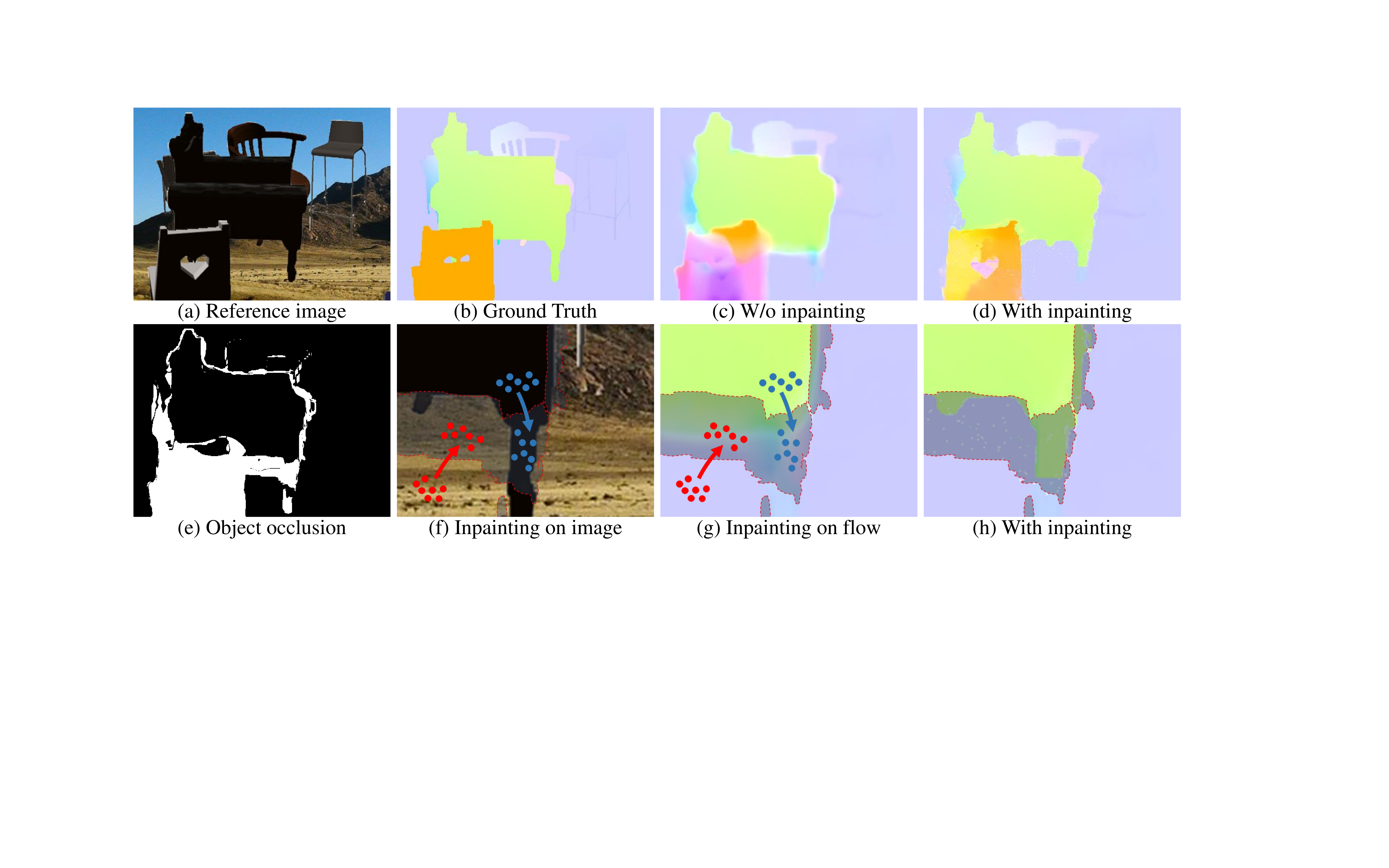}\\
  \vspace{-2mm}
  \caption{(a) reference image. (b) ground truth flow. (c)(d) results without / with the optical flow refinement. (e) detected occlusion map. (f) a zoom-in window with mask (dark region) overlaid on the image. Here an appearance flow is learned on the image domain, which is further used to inpaint the occluded regions by the non-occluded ones. (g) the appearance flow of (f) is applied to guide the propagation in the flow field. (h) The propagated results in the zoom-in region.}\label{fig:intro}
  \vspace{-4mm}
\end{figure}

To this end, we propose our OccInpFlow which not only locates but also inpaints real occlusions.
We modified the appearance flow network that originally proposed for the task of image inpainting~\cite{Ren_2019} to inpaint missing flows contextually. Fig.~\ref{fig:intro} demonstrates our motivation. Fig.~\ref{fig:intro} (a), (b) and (e) show examples of reference frame, groundtruth flow and detected object occlusion mask, respectively. Fig.~\ref{fig:intro} (c) simply discards occluded regions during training. The flow on the back of the chair is incorrect. Fig.~\ref{fig:intro} (d) is our result with occlusion regions inpainted by our appearance flow refinement block. The motion of the chair is now correct. Specifically, we show a zoom-in region in Fig.~\ref{fig:intro} (f), where the inpainting is applied on the image domain to inpaint from non-occlusion regions to occlusion regions according to the image contents (red and blue arrows). The same inpainting procedure is copied to the flow field (Fig.~\ref{fig:intro} (g)). Fig.~\ref{fig:intro} (h) shows the inpainted flow field.

In addition, we further divide occlusions into two types, object occlusion and boundary occlusion. For the former one, which occurs within the image, we inpaint according to contextual relevance. For the later one, which is caused by displacement beyond the image boundary, we propose a boundary dilated warp to correct its loss. Moreover, our network backbone is built upon the supervised framework IRR-PWC~\cite{irrpwc} by absorbing its merits but changed from supervised to unsupervised setting.

Through extensive experiments on multiple representative data sets, we demonstrate that our OccInpFlow can significantly improve the performance, exceeding all existing unsupervised methods. In particular, on MPI-Sintel benchmark test, we achieve EPE = $5.79$ on the Clean pass (the second best method is $6.18$) and EPE = $7.28$ on the Final pass (the second best method is $7.40$), which even outperform some supervised methods. Our code and trained models will be available online. Our contribution are summarized as follows:
\vspace{-1mm}
\begin{itemize}
    \item An occlusion inpainting framework named \textbf{OccInpFlow} is proposed for unsupervised optical flow estimation.
\vspace{-1mm}
    \item A new appearance-flow network named \textbf{AppFlowNet-Lite} is proposed to inpaint occluded flows based on the image content to provide additional guidance in occlusions. Such important regions are either treated as non-occlusions or totally abandoned previously.
\vspace{-1mm}
    \item A \textbf{Boundary Dilated Warping} is proposed to handle pixels caused by frame boundary occlusions, which is effective and not attempted before.
\end{itemize}

\section{Related Work}\label{sec:related_work}\vspace{-2mm}


\paragraph{Supervised deep optical flow.}
Supervised methods learn optical flow from annotated data~\cite{cnn_patch_match2017,zhao2020maskflownet,CostVolume_cvpr2017,VolumetricCN2019}.
FlowNet~\cite{Flownet_flyingchairs} is the first work to estimation optical flow using a fully convolutional network.
The following work FlowNet2~\cite{FlowNet2} extends FlowNet by stacking networks iteratively, which largely improves the performance.
For large displacements, SPyNet~\cite{spynet2017} proposed to warp images based on the coarse-to-fine manner.
Then, PWCNet~\cite{pwc_net,pwc_net_tpami} and LiteNet~\cite{LiteFlowNet,Lightweight_tpami} proposed to calculate cost volume after warping features, resulting in efficient and lightweight networks.
Recently, IRRNet~\cite{irrpwc} proposed an iterative residual refinement scheme for network design, which achieves state-of-the-art result.
However, The performance of these deep learning based methods depend heavily on the training data, which is prohibitively expensive in practical applications. \vspace{-1mm}

\paragraph{Unsupervised deep optical flow.} 
Unsupervised methods do not require flow annotations~\cite{unsup_ICIP_2016,Ren2017aaai,Jason2016}, where the photometric loss is optimized during training. However, the photometric loss cannot work in occlusion regions and even lead to incorrect guidance.
Therefore, works~\cite{wang2018,unflow_2018aaai} proposed to exclude occlusion regions from photometric loss for performance improvement.
Recently, researchers introduced multi-frame formulation~\cite{unflow_multi_occ}, Epipolar constrain~\cite{Epipolar_flow_2019cvpr}, depth estimation \cite{Anurag2019,dfnet_zou_2018,Yin2018GeoNetUL,Liu2019UnsupervisedLO}, and stereo matching \cite{Wang2019UnOSUU} to further constraint the problem.
DDFlow~\cite{Pengpeng2019} and SelFlow~\cite{Liu2019CVPR} proposed to learn optical flow based on data distillation. They artificially create occlusions for self supervision. Although the performance can be largely improved, the real occlusions still have no guidance.
In this paper, we provide guidance in occlusion regions by our appearance flow warp and boundary warp. \vspace{-1mm}



\paragraph{Appearance flow.} 
Appearance flow is first proposed by~\cite{first_appearance_flow_for_view_synthesis} for view synthesis. Appearance flow can copy pixels from source images for target object generation, which is superior than generating from scratch. Recently, appearance flow is applied for image inpainting~\cite{Ren_2019}, where the appearance flow can propagate features from existing regions to missing regions for texture generation. In this paper, we modify the appearance flow for optical flow inpainting.  \vspace{-1mm}
\section{Algorithm}\label{sec:algo}\vspace{-2mm}






\subsection{Network Structure}
\begin{figure*}
  \centering
  \includegraphics[width=0.97\linewidth]{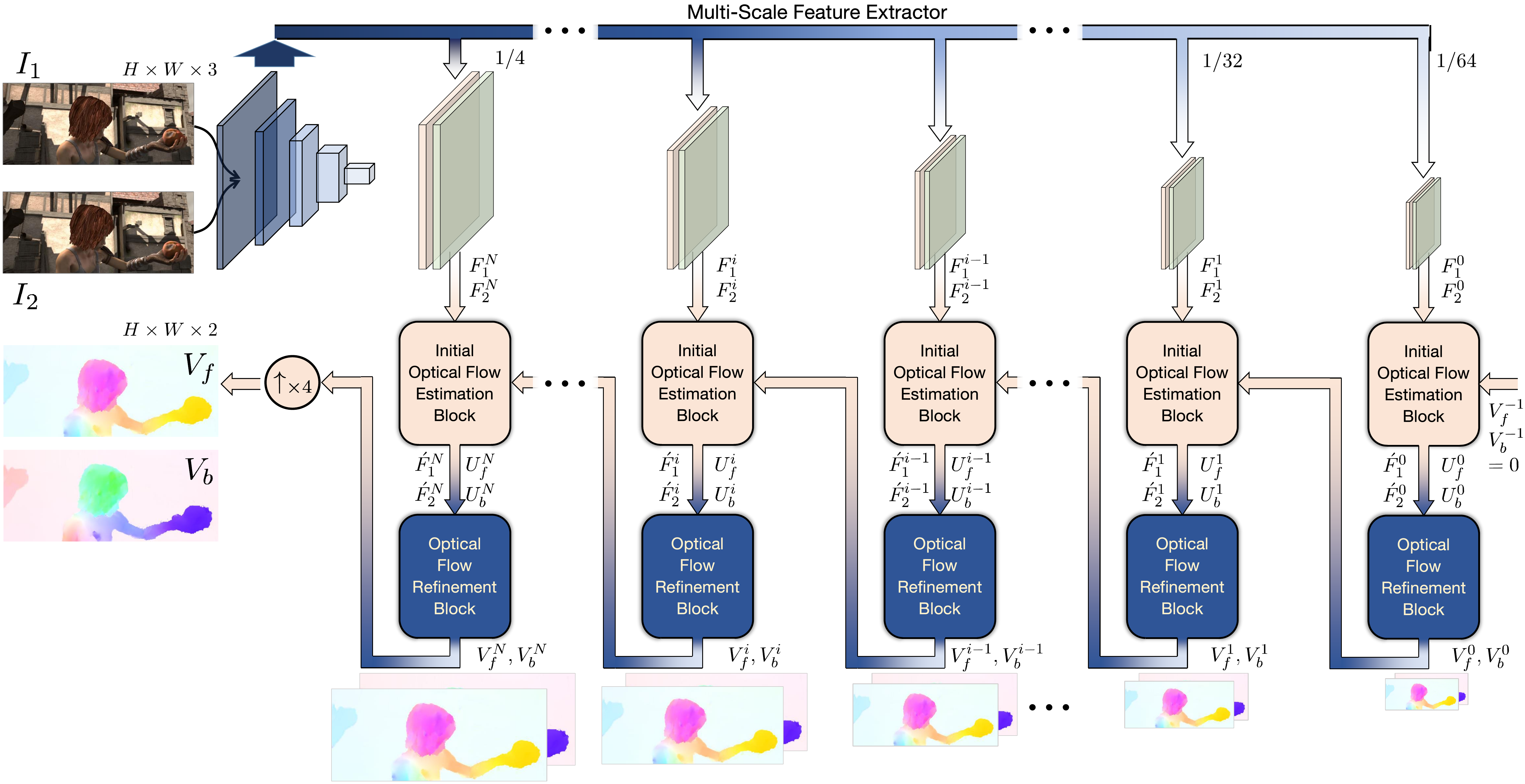}\vspace{-2mm}
  \caption{The network structure of our OccInpFlow Network, which involves 3 main parts, a multi-scale feature extractor, and an initial optical flow estimation block (Fig.~\ref{subfig:estimation-block}) together with a refinement block (Fig.~\ref{subfig:refinement-block}) working on each scale.}\label{fig:network-structure}\vspace{-3mm}
\end{figure*}

\vspace{-1mm}
Our method is built upon a supervised optical flow estimation framework IRR-PWC~\cite{irrpwc}, which is further adapted to enable unsupervised learning and occlusion awareness. It takes two frames $I_1$ and $I_2$ of size $H\times W\times 3$ as input, and produces forward and backward optical flows $V_f, V_b$ of size $H\times W\times 2$ as output. The entire network can be divided into three parts, i.e., a feature extractor that extracts multi-scale feature maps of inputs, an initial optical flow estimation block and a newly introduced optical flow refinement block. We illustrate the network structure in Fig.~\ref{fig:network-structure}.


\subsubsection{Multi-Scale Feature Extractor}\vspace{-1mm}
Given two consecutive frames $I_1, I_2$, we first extract $(N + 1)$ scales of feature maps $\{F^i_k\}, i = 0,1,...,N$ for each $I_k~(k \in \{1,2\})$ separately, by a fully convolutional encoder $\mathcal{E}$. For each frame, its successive feature maps $F^i_k$ and $F^{i-1}_k (k \in \{1,2\})$ follow a relationship of $2\times$ scale. In our design, we denote the feature map of $1/64$ image size as the $0$-th one, so that the $i$-th feature map is of $2^{i-6}$ image size, $i = 0,1,...,N$. As such, the feature maps extracted are written as follows.
\vspace{-0.5mm}
\begin{align}\label{eq:multi-scale-feature-extractor}
\big\{F^i_k\big\}= \mathcal{E}(I_k), \quad \text{for}~i \in \{0, 1, ..., N\}~\text{and}~k \in \{1, 2\}
\end{align}
In practice, we set $N$ to $4$ so that the finest scaled feature maps extracted by $\mathcal{E}$ are of $1/4$ image size. This size is a balanced result considering the accuracy and efficiency.

\subsubsection{Initial Optical Flow Estimation Block}\vspace{-1mm}
We further generate \textit{initial} forward and backward optical flows for each scale in a coarse-to-fine manner. For the $i$-th scale, the feature pair $(F^i_1, F^i_2)$ is further fed to an estimator $\mathcal{D}$. Together with the \textit{refined} optical flows in the $(i-1)$-th scale, i.e. $V^{i-1}_f, V^{i-1}_b$, the estimator $\mathcal{D}$ produces two initial optical flows $U^i_f, U^i_b$ as follows. 
\begin{align}\label{eq:initial-optical-flow-estimator}
U^i_f = \mathcal{D}\big(F^i_1, F^i_2, V^{i-1}_f\big), \quad
U^i_b = \mathcal{D}\big(F^i_2, F^i_1, V^{i-1}_b\big)
\end{align}
%
Specifically, the feature maps $\{F^i_1\}$ from multiple scales may have various numbers of channels, we add one $1\times1$ conv layer following $F^i_1$, to get a feature map $\acute{F}^i_1$ of a uniform number of channels $N_c$ (Eq.~\ref{eq:initial-optical-flow-estimator-1}). Second, as the refined optical flow $V^{i-1}_f$ from the $(i-1)$-th scale is only $1/2$ size of $F^i_k$, we upsample it to obtain its $2\times$ version $\hat{V}^{i-1}_f$, which is further applied to warp $F^i_2$ to obtain $\tilde{F}^i_2$ (Eq.~\ref{eq:initial-optical-flow-estimator-2}). We further calculate the correlation of $F^i_1$ and $\tilde{F}^i_2$ (Eq.~\ref{eq:initial-optical-flow-estimator-3}), and finally feed the three tensors to a flow estimator followed by a context network, to obtain the initial forward optical flow $U^i_f$ (Eq.~\ref{eq:initial-optical-flow-estimator-4}). To sum up, the process above is described equationally as follows. \vspace{-2mm}
%
\begin{minipage}[c]{.4\linewidth}
\begin{equation}
\acute{F}^i_1 = \text{Conv}_{1\times1\times N_c}(F^i_1) \label{eq:initial-optical-flow-estimator-1}
\end{equation}
\end{minipage}
\begin{minipage}[b]{.6\linewidth}
\begin{equation}
\hat{V}^{i-1}_f = \text{Sample}_{\uparrow_{\times2}}(V^{i-1}_f), \quad \tilde{F}^i_2 = \mathcal{W}(F^i_2, \hat{V}^{i-1}_f) \label{eq:initial-optical-flow-estimator-2}
\end{equation}
\end{minipage}

\begin{figure*}
  \centering
  \subfigure[Initial Optical Flow Estimation Block.]{
  \includegraphics[width=0.46\linewidth]{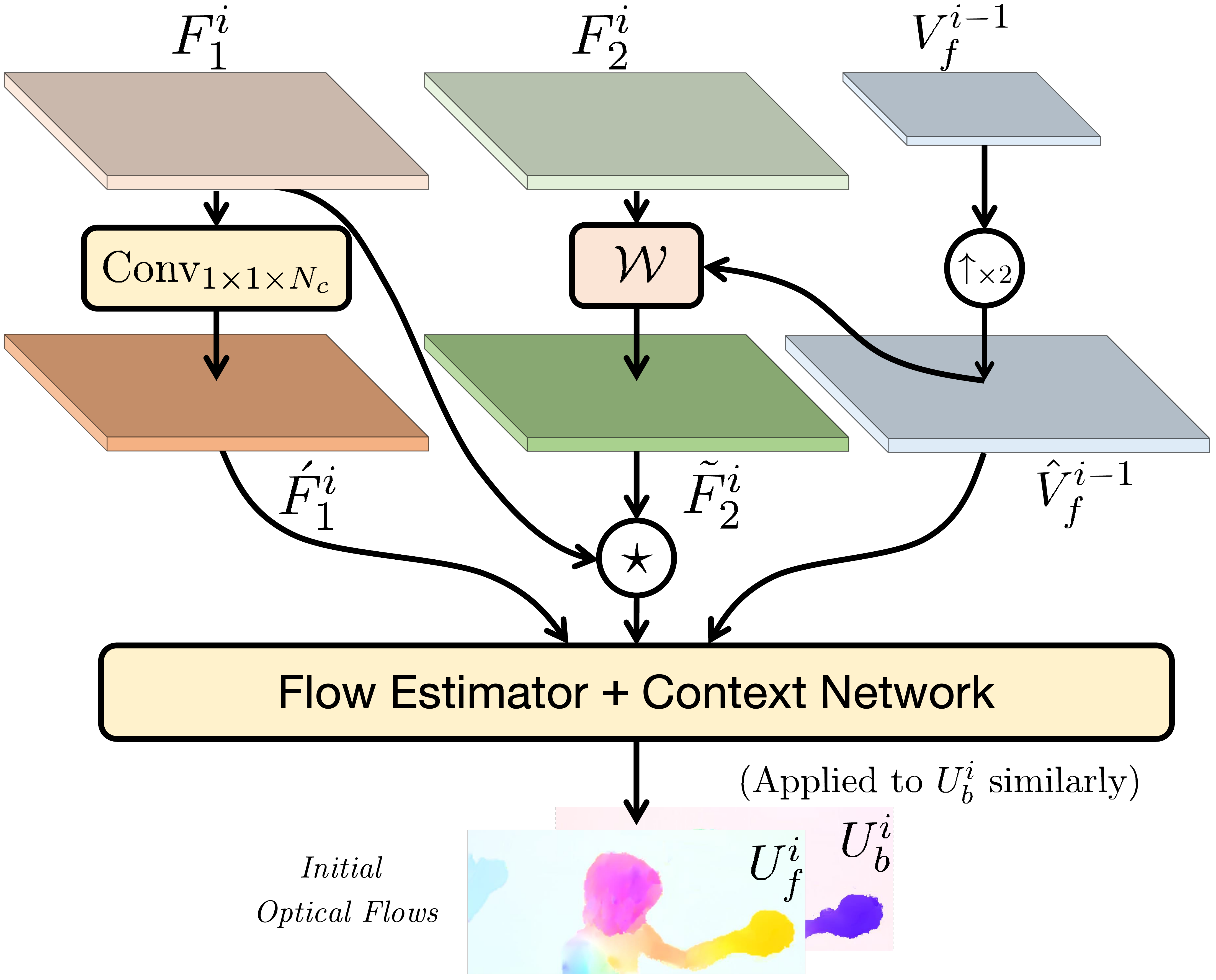}\label{subfig:estimation-block}
  }
  \hfill
  \subfigure[Optical Flow Refinement Block.]{
  \includegraphics[width=0.46\linewidth]{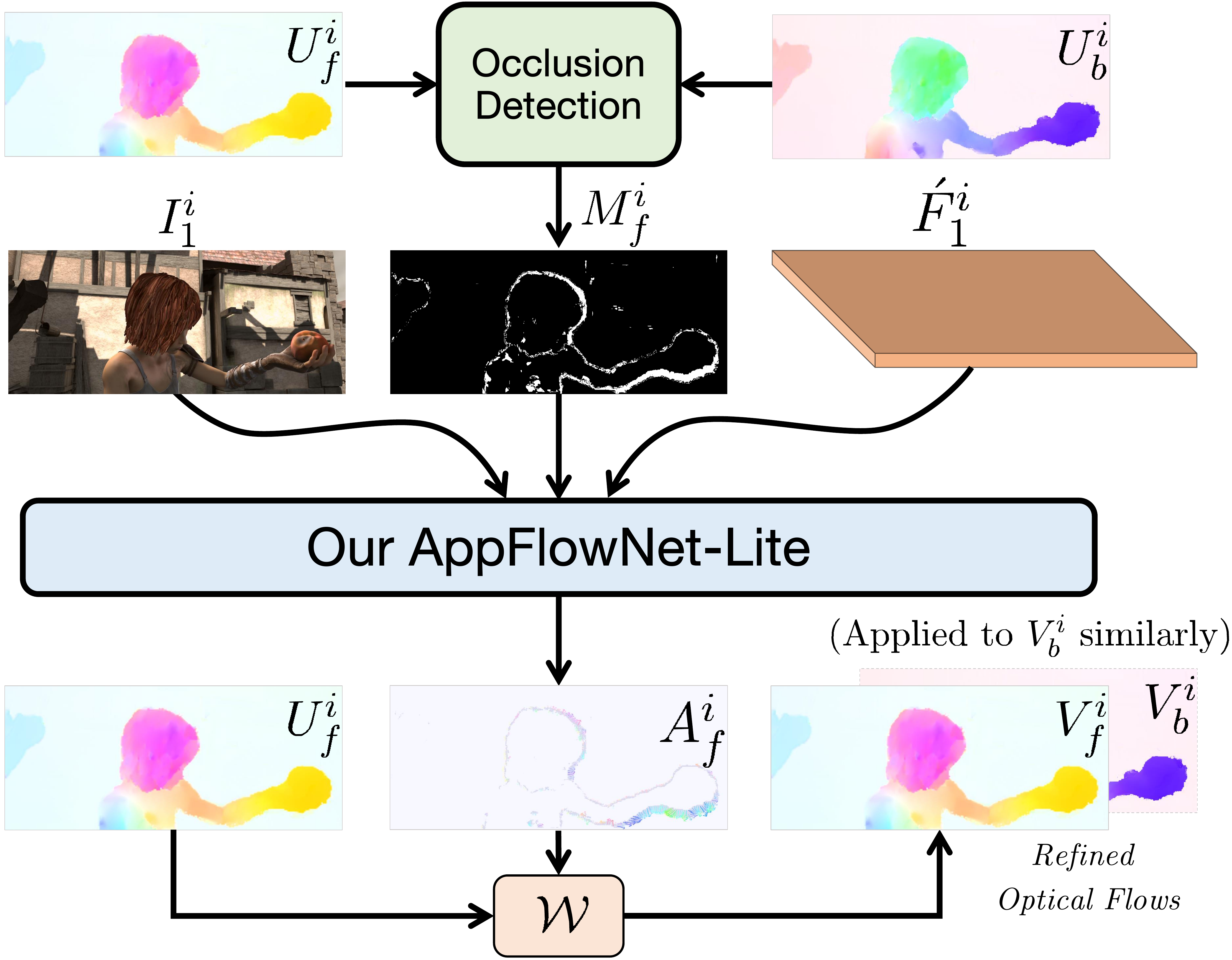}\label{subfig:refinement-block}
  }
  \vspace{-2mm}
  \caption{The detailed structures of the initial optical flow estimation block (a) and the optical flow refinement block (b). Due to the limited size of the paper, please refer to the supplementary material for the structure of our AppFlowNet-Lite.}\label{fig:estimation-block-and-refinement-block}\vspace{-3mm}
\end{figure*}

\vspace{1mm}
\begin{minipage}[c]{.4\linewidth}
\begin{equation}
\Phi_i = F^i_1 \star \tilde{F}^i_2~~~~~~~~~~~~~~~~\label{eq:initial-optical-flow-estimator-3}
\end{equation}
\end{minipage}
\begin{minipage}[b]{.6\linewidth}
\begin{equation}
U^i_f = \mathcal{D}\big(F^i_1, F^i_2, V^{i-1}_f\big) = \mathcal{H}(\acute{F}^i_1, \Phi_i, \hat{V}^{i-1}_f)~~~~~ \label{eq:initial-optical-flow-estimator-4}
\end{equation}
\end{minipage}\vspace{2mm}

%
%
where $\mathcal{W}$ is warping operation achieved by pixel re-mapping, $\star$ is the correlation operator and $\mathcal{H}$ represents the combination of the flow estimator and context network. Similarly, we obtain the backward optical flow $U^i_b$ by swapping $F^i_1, F^i_2$ and replacing $V^{i-1}_f$ by $V^{i-1}_b$, i.e. $U^i_b = \mathcal{D}(F^i_2, F^i_1, V^{i-1}_b)$. The operations above apply for all scales except $i=0$, where we set $V^{-1}_f = V^{-1}_b = 0$. A detailed structure of $\mathcal{D}$ is illustrated in Fig.~\ref{fig:estimation-block-and-refinement-block}.

\subsubsection{Optical Flow Refinement Block}\vspace{-1mm}
The initial optical flows estimated above commonly have incorrect pixels caused by occlusions. Severe artifacts may occur if we directly apply warping. We propose to first detect an occlusion map and then to inpaint the occluded regions using the optical flow in the non-occluded ones. In our framework, we obtain the occlusion map $M^i_f$ by comparing the initial forward and background optical flows $U^i_f, U^i_b$, and design a light version of the Appearance Flow Network~\cite{Ren_2019} (AppFlowNet) as our inpainting approach, named \textbf{AppFlowNet-Lite} (detailed later). Specifically, we follow the forward-backward consistency prior of optical flow as in~\cite{unflow_2018aaai} to check whether a pixel is still within its neighborhood after being moved by the forward and backward optical flow vectors, i.e.
\begin{align}\label{eq:occlusion-mask-check}
M^i_f(\bm{p}) =
\begin{cases}
  1, & \mbox{if~} \big|V^i_f(\bm{p}) + V^i_b\big(\bm{p} + V^i_f(\bm{p})\big)\big| > \alpha_1\big(|V^i_f(\bm{p})| + |V^i_b(\bm{p})|\big) + \alpha_2 \\
  0, & \mbox{otherwise}.
\end{cases}
\end{align}
where $\bm{p}$ is a pixel coordinate and $\alpha_1, \alpha_2$ are hyper-parameters to control the neighborhood size. After $M^i_f$ is acquired, we feed $\acute{F}^i_1, I^i_1, M^i_f$ to the AppFlowNet-Lite $\mathcal{A}$ to produce an appearance flow $A^i_f$. In $A^i_f$, the matching vectors are generated for the pixels in occlusions, so that by warping $U^i_f$ with $A^i_f$, a refined flow $V^i_f$ is achieved. Likewise, $V^i_b, A^i_b$ and $M^i_b$ can be obtained similarly. In summary,
\begin{equation}\label{eq:optical-flow-refinement}
V^i_\beta = \mathcal{W}(U^i_\beta, A^i_\beta), \quad A^i_\beta = \mathcal{A}(\acute{F}^i_k, I^i_k, M^i_\beta), \quad \langle k, \beta\rangle \in \{ \langle 1, f\rangle, \langle 2, b\rangle  \}
\end{equation}
where $I^i_k$ are the corresponding down-sampled version of $I_k, k\in\{1,2\}$ in $i$-th scale. Note that here we also feed $\mathcal{A}$ with $\acute{F}^i_k$ just for avoiding redundant feature extraction steps and saving network parameters. See Fig.~\ref{fig:estimation-block-and-refinement-block} for the illustration of this block.
\vspace{-1mm}

\paragraph{Our AppFlowNet-Lite.}
The original AppFlowNet~\cite{Ren_2019} follows a classical Encoder-Decoder framework and involves a moderate scale of parameters. To facilitate the efficiency of our framework, we follow the similar idea and design a light version of it. We remove several layers from the original network to make our implementation handle only 2 scale of feature maps instead of 4 scales. Also we make the parameters of the two decoding layers shared so that more weights can be saved. With these schemes, our AppFlowNet-Lite contains only \textbf{3.82}\% the parameters of the original one (2.06 M vs 53.86 M), but the performance is demonstrated stable enough. We show the structure AppFlowNet-Lite in our supplementary materials.


\vspace{1mm}

With the scale becoming finer, we obtain higher quality flows with occlusions well inpainted. After we finish the $N$-th scale with $V^N_f$ and $V^N_b$, we directly upsample them to the size of $H\times W\times 2$, and apply the refinement block to them. By doing this, $V_f, V_b$ are generated as the final output. 

\subsection{Unsupervised Training}
We customize an unsupervised loss to characterize the pixel alignment. Due to the existence of two types of occlusion, that are caused by internal objects and the limited image plane, we design corresponding warping strategies. For the former one, we trivially utilize the standard warping operation $\mathcal{W}$, while for the latter case, we introduce a Boundary Dilated Warping (Section~\ref{subsec:bdw}) before formulating our unsupervised loss as detailed in Section~\ref{subsec:unsupervised-loss}.

\subsubsection{Boundary Dilated Warping}\label{subsec:bdw}
After computing the occlusion mask $M^i_f, M^i_b$ defined in Eq.~\ref{eq:occlusion-mask-check}, we further calculate the in-frame occlusion maps $M^i_{\Omega, f}, M^i_{\Omega, b}$ by checking whether a pixel would be moved outside the image plane by the optical flow, i.e.
\begin{align}\label{eq:in-frame-regions}
M^i_{\Omega, \beta}(\bm{p}) =
\begin{cases}
  M^i_\beta(\bm{p}), & \mbox{if~} \bm{p} + V^i_\beta(\bm{p}) \in \Omega\\
  0, & \mbox{otherwise}.
\end{cases}
~~,\quad\beta \in \{f, b\}
\end{align}
where $\Omega$ is the image plane. In other words, the regions where $M^i_{\Omega, \kappa}(\bm{p}) = 1$ represent the internal occlusion caused by objects instead of being limited by the image size. For the rest of the pixels, some of them may be moved outside the image plane. This type of pixels usually stay close to the four external image boundaries. To overcome this issue, we propose a boundary dilated warping strategy $\mathcal{W}_\Omega$ for the regions where $M^i_{\Omega, \kappa}(\bm{p}) = 0$.

Specifically, in the training stage, we actually feed the network by a cropped version of the raw image pairs which denoted as $I^r_1, I^r_2$, instead of themselves. That is to say, $I_1, I_2$ are achieved as follows,
\begin{align}\label{eq:crop}
I_k(\bm{p}) = I^r_k(\bm{p}_0 + \bm{p}), \quad k \in \{1, 2\}
\end{align}
where $\bm{p}_0$ is the upper-left coordinate of the cropped region in $I^r_k, k\in\{1,2\}$, and we need to keep $I_k$ is totally covered by $I^r_k$. It looks like $\{I^r_k\}$ are the ``dilated'' version of $\{I_k\}$. As such, our boundary dilated warping is expressed as below,
\begin{equation}\label{eq:bdw}
\tilde{I}_k(\bm{p}) = \mathcal{W}_\Omega(I^r_k, V_\beta, \bm{p}_0)(\bm{p}) = I^r_k\big(\bm{p}_0 + \bm{p} + V_\beta(\bm{p})\big), \quad\langle k, \beta\rangle \in \{ \langle 1, f\rangle, \langle 2, b\rangle  \}
\end{equation}

\subsubsection{Unsupervised Loss}\label{subsec:unsupervised-loss}
\paragraph{1) Alignment losses.}
For unsupervised optical flow estimation tasks, the core metric is the alignment quality. Here we follow the popular brightness constancy assumption that the two frames should have similar pixel intensities so as to utilize the \textbf{photometric loss} $\mathcal{L}_p$ as the main error metric. However, considering aligned pixels do not exist in the internal occluded regions, we only evaluate $\mathcal{L}_p$ on the non-occluded area. As mentioned in Section~\ref{subsec:bdw}, we apply Boundary Dilated Warping $\mathcal{W}_\Omega$ to each frame instead of the standard warping $\mathcal{W}$. As a result, $\mathcal{L}_p$ is formulated as follows.
\begin{align}\label{eq:photo-loss}
\mathcal{L}_p=\frac{\sum_{\bm{p}}{\Big[\psi(I_1-\tilde{I}_2)\odot (1-M_f)\Big]}}{\sum_{\bm{p}}{(1-M_f)}}+\frac{\sum_{\bm{p}}{\Big[\psi(I_2-\tilde{I}_1)\odot (1-M_b)\Big]}}{\sum_{\bm{p}}{(1-M_b)}}
\end{align}
where $\psi(x)=(|x|+\epsilon)^q$ is a robust loss function with $\epsilon = 0.01$ and $q = 0.4$ in our experiments, and $\odot$ is element-wise multiplier.

To robustly handle the case that brightness constancy assumption is not met, we further apply census transform~\cite{unflow_2018aaai} $\mathcal{C}(\cdot)$ to each of images $I_1,I_2,\tilde{I}_1,\tilde{I}_2$ before the calculation. Census transform is a local binary pattern encoding the local pixel intensity ordering, so that it has a certain capability to overcome the illuminance variation. To substitute $I_1,I_2,\tilde{I}_1,\tilde{I}_2$ by their census transformed versions $\mathcal{C}(I_1),\mathcal{C}(I_2),\mathcal{C}(\tilde{I}_1),\mathcal{C}(\tilde{I}_2)$, we obtain a \textbf{census loss} $\mathcal{L}_c$ of similar equation to $\mathcal{L}_p$ as in Eq.~\ref{eq:photo-loss}. \vspace{-3mm}

\paragraph{2) Inpainting loss of AppFlowNet-Lite.}
To minimize the photometric and census losses defined in Eq.~\ref{eq:photo-loss}, high-quality refined optical flows $V^i_f, V^i_b$ play a key role because they explicitly influence the warping results by the boundary dilated warping operation $\mathcal{W}_\Omega$. And then high-quality appearance flows $A^i_f, A^i_b$ are required, so that the AppFlowNet-Lite embedded in the optical flow refinement block needs to be well trained. For this reason, we formulate an inpainting loss:
\begin{align}\label{eq:inpainting-loss}
\mathcal{L}_a= \sum_{i=0}^N{\Bigg(
\frac{\sum_{\bm{p}}{\Big[\psi(I_1^i-\tilde{Q}^i_1)\odot M_f^i\Big]}}{\sum_{\bm{p}}{M_f^i}}  +   \frac{\sum_{\bm{p}}{\Big[\psi(I_2^i-\tilde{Q}^i_2)\odot M_b^i\Big]}}{\sum_{\bm{p}}{M_b^i}}
\Bigg)
}
\end{align}
where $\tilde{Q}^i_1 = \mathcal{W}(Q_1^i, A_b^i), \tilde{Q}^i_2 = \mathcal{W}(Q_2^i, A_f^i)$, and here $Q^i_k, k\in\{1,2\}$ are the masked versions of $I^i_k$ by the occlusion maps $M^i_f, M^i_b$. That is to say, $Q^i_k(\bm{p}) = I^i_k(\bm{p}) \cdot \big(1 - M^i_\beta(\bm{p})\big), \text{for~} \langle k, \beta\rangle \in \{\langle1,f\rangle, \langle2,b\rangle\}$. $N$ is the index of the finest scale, being set to $5$ in our experiments. Eq.~\ref{eq:inpainting-loss} reveals that the reconstruction error between the inpainted image and itself is evaluated on the internal occluded regions only. By minimizing this loss, our AppFlowNet-Lite well learns how to proceed ``PatchMatch''-like hole filling for the RGB image, so as to be further applied to the optical flows for refinement. Considering there is no out-of-boundary problem, we just use a standard warping $\mathcal{W}$. \vspace{-3mm}

\paragraph{3) Smoothness loss.}
We also add a regularization term to constrain the smoothness of the generated optical flows following the idea in~\cite{Jason2016}. That is to minimize the $l_1$ loss of the first order gradient of the predicted optical flows, i.e.
\begin{align}\label{equ:smoothness_loss}
\mathcal{L}_s=|\nabla{V_f}|+|\nabla{V_b}|
\end{align}
Eventually, our unsupervised loss $\mathcal{L}$ is defined as the combination of the 4 losses above,
\begin{align}\label{equ:total_loss}
\mathcal{L}=\lambda_p\mathcal{L}_p+\lambda_c\mathcal{L}_c+\lambda_a\mathcal{L}_a+\lambda_s\mathcal{L}_s
\end{align}
We set $\lambda_p=1$, $\lambda_c=0.01$, $\lambda_a=1$ and $\lambda_s=0.05$ in our experiments.

\section{Experimental Results}\label{sec:results}
\subsection{Dataset and Implementation Details} \vspace{-2mm}
\paragraph{Dataset.} We conduct our experiments and comparison on the following 3 datasets. \vspace{-2mm}
\begin{enumerate}[1)]
  \item \textbf{Flying Chairs.}  Flying chairs~\cite{Flownet_flyingchairs} is a synthetic data set created by randomly moving chair images, which contains $22,872$ pairs for training and $640$ pairs for testing. Following DDFlow~\cite{Pengpeng2019}, the training images are used for unsupervised training (no groundtruth), and the test images are used for validation. The image size is $384\times480$ and we crop patches with $320\times320$ during training.
  \vspace{-1mm}
  \item \textbf{KITTI.} KITTI2012~\cite{KITTI_2012} contains $194$ training pairs and $195$ test pairs, and KITTI2015~\cite{KITTI_2015} contains $200$ training pairs and $200$ test pairs. Two data sets also contain multi-view extension data sets with no groundtruth. Following~\cite{Ren2017aaai, Pengpeng2019, Liu2019CVPR}, we use $13,372$ image pairs of the multi-view extension for training, and use the train set of KITTI2012 and KITTI2012 for validation. Patches are cropped to $320\times896$. Results are uploaded to KITTI website for benchmark comparison.
  \vspace{-1mm}
  \item \textbf{MPI-Sintel.} MPI-Sintel \cite{Butler2012} is a synthetic data set. It contains $1,041$ training pairs and $564$ test pairs. Two different sets are provided, `Clean' and `Final'. We use $1,041$ pairs from `Clean' for training. Images are cropped to $320\times768$. We also upload results of test set for online evaluation.
\end{enumerate}\vspace{-3mm}

\paragraph{Implementation Details.}

We developed our code with PyTorch, and the whole training was completed in $500k$ iterations which costs about 48 hours by 2 NVIDIA GeForce GTX 2080Ti GPUs. The total number of parameters of our network is around 5.26M. To stabilize the training, we divide the training process into 3 stages. First, we set the learning rate to $10^{-4}$ to learn initial optical flows with AppFlowNet-Lite disabled and without occlusion handling. After about $300k$ iterations, we reduce the learning rate to $10^{-5}$ and exclude the occlusion regions in $\mathcal{L}_p$ to learn accurate optical flows in the non-occluded regions for $100k$ iterations. Finally, our AppFlowNet-Lite is enabled and predictions from the non-occluded regions is propagated to the occluded ones. We use the average endpoint error (EPE) to evaluate the quality of our predicted optical flow. The percentage of erroneous pixels (F1) is also used as evaluation metric on the test set of KITTI2015. \vspace{-1mm}

\subsection{Comparison with Existing Methods} 
\begin{figure}[t]
  \centering
  \includegraphics[width=0.99\textwidth]{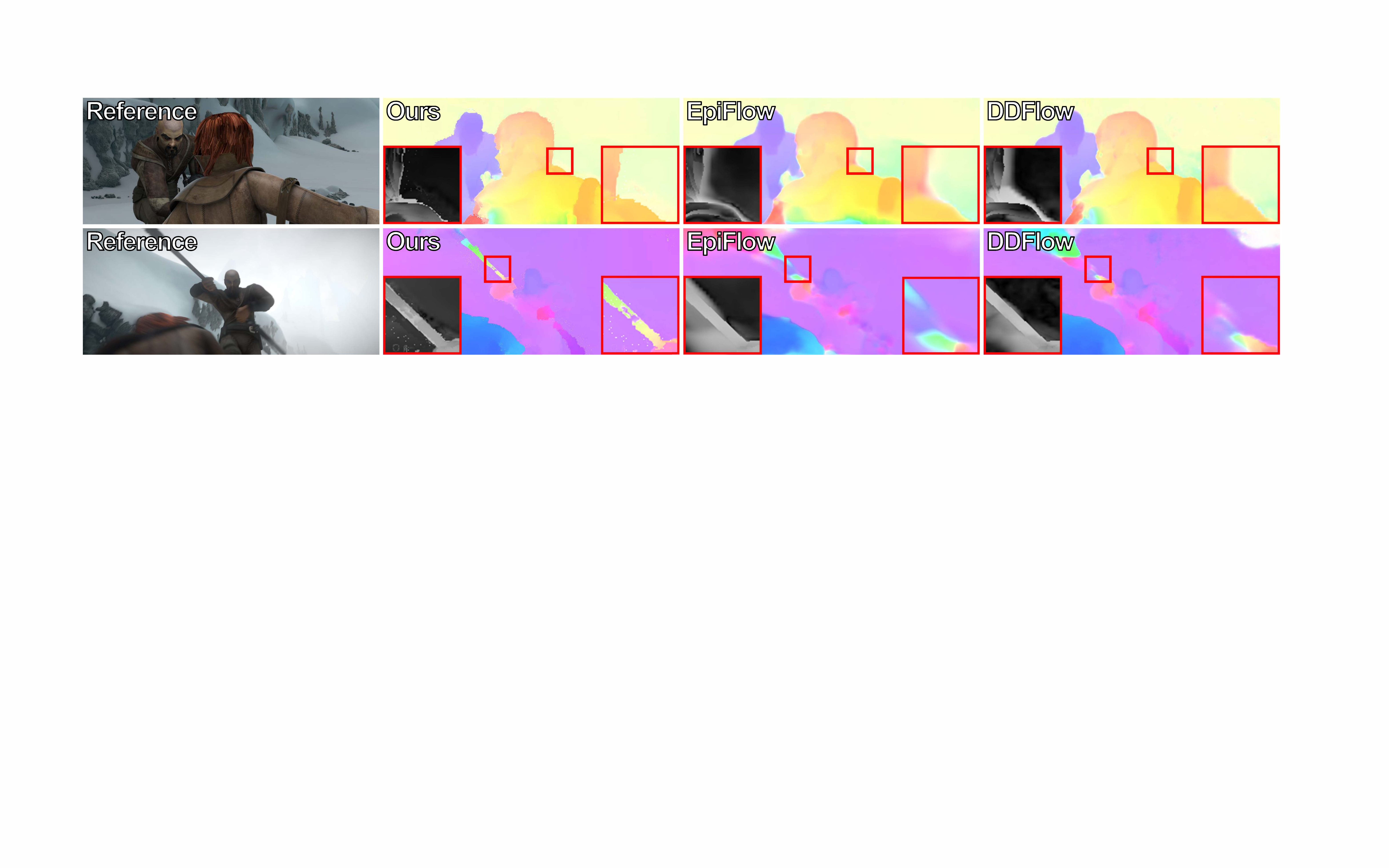}\\
  \vspace{-2mm}
  \caption{Qualitive results on MPI-Sintel benchmark. Results from Sintel Clean(top), and Sintel Final(bottom) are shown. We compare our method with DDFlow \cite{Pengpeng2019} and EpiFlow \cite{Epipolar_flow_2019cvpr}. The zoom in flows (right bottom) and the zoom in error maps (left bottom) of each sample are shown. }\label{fig:qualitive_results_on_MPI_sintel}\vspace{-2mm}
\end{figure}

\vspace{-2mm}
As shown in Table~\ref{comparision_sota}, we compare the proposed approach with previous representative works on Flying Chairs, KITTI and Sintel data sets. The proposed approach outperforms all the previous works on all the data sets. On Flying Chairs, we decrease the EPE value from previous state-of-the-art $2.97$ to $2.62$, and also outperform some fully supervised model such as FlowNetS~\cite{Flownet_flyingchairs} and SpyNet~\cite{spynet2017}. On KITTI2012 online evaluation, we improve the EPE value from $3.0$ of DDFlow~\cite{Pengpeng2019} to $2.1$ with $30\%$ improvement. On KITTI2015 train set, we improve the EPE value from $5.55$ of EpiFlow~\cite{EpicFlow_2015} to $4.57$. On the test benchmark of MPI-Sintel, we achieve $5.79$ on the clean pass and $7.28$ on the final pass, which both are better than all the previous unsupervised methods. Some qualitative results are shown in Fig.~\ref{fig:qualitive_results_on_MPI_sintel}. 

\begin{table}[!t]
\centering
\resizebox*{\textwidth}{!}{
\begin{tabular}{
>{\centering\arraybackslash}p{0.3cm}
p{3cm} 
>{\centering\arraybackslash}p{1.2cm}
>{\centering\arraybackslash}p{1.2cm}
>{\centering\arraybackslash}p{1.2cm}
>{\centering\arraybackslash}p{1.2cm}
>{\centering\arraybackslash}p{1.8cm}
>{\centering\arraybackslash}p{1.2cm}
>{\centering\arraybackslash}p{1.2cm}
>{\centering\arraybackslash}p{1.2cm}
>{\centering\arraybackslash}p{1.2cm}
}
\toprule
\multicolumn{2}{c}{\multirow{2}{*}{Method}} & Chairs &\multicolumn{2}{c}{KITTI 2012} & \multicolumn{2}{c}{KITTI 2015} & \multicolumn{2}{c}{Sintel Clean}&\multicolumn{2}{c}{Sintel Final}\\
\cmidrule(lr){3-3}\cmidrule(lr){4-5} \cmidrule(lr){6-7} \cmidrule(lr){8-9} \cmidrule(lr){10-11}
&&test&train & test &train & test (F1-all) &train &test &train &test
\\
\midrule
\multirow{9}{*}{\rotatebox{90}{Supervised}}
& FlowNetS\cite{Flownet_flyingchairs}    &2.71  &8.26  &  --  &  --  &   --  & 4.50 &7.42  &5.45  &8.43  \\
& FlowNetS+ft\cite{Flownet_flyingchairs} &  --  &7.52  &9.1   &  --  &   --  &(3.66)&6.96  &(4.44)&7.76  \\
& SpyNet\cite{spynet2017}                    &2.63  &9.12  &  --  &  --  &   --  &4.12  &6.69  &5.57  &8.43  \\
& SpyNet+ft\cite{spynet2017}                 &  --  &8.25  &10.1  &  --  &35.07\%&(3.17)&6.64  &(4.32)&8.36  \\
& FlowNet2\cite{FlowNet2}                &  --  &4.09  &  --  &10.06 &   --  & 2.02 & 3.96 & 3.14 &6.02  \\
& FlowNet2+ft\cite{FlowNet2}             &  --  &(1.28)& 1.8  &(2.3) &11.48\%&(1.45)& 4.16 &(2.01)&5.74  \\
& PWC-Net\cite{pwc_net}                  & 2.00 & 4.57 &  --  &13.20 &   --  & 3.33 &  --  & 4.59 &  --  \\
& PWC-Net+ft\cite{pwc_net}               &  --  &(1.45)& 1.7  &(2.16)&9.60\% &(1.70)& 3.86 &(2.21)&5.13  \\
& IRR-Net\cite{irrpwc}                   & 1.67 &  --  &  --  &(1.63)&7.65\% &(1.92)& 3.84 &(2.51)&4.58  \\
\midrule
\multirow{11}{*}{\rotatebox{90}{Unsupervised}}
& BackToBasic+ft\cite{Jason2016}            & 5.3  & 11.3 & 9.9  &  --  &   --  &  --  &  --  &  --  &  --  \\
& DSTFlow+ft\cite{Ren2017aaai}     & 5.11 & 10.43& 12.4 & 16.79& 39\%  &(6.16)&10.41 &(6.81)&11.27 \\
& UnFlow-CSS+ft\cite{unflow_2018aaai}      &  --  & 3.29 &  --  & 8.10 & 23.3\%&  --  &9.38  &(7.91)&10.22 \\
& OAFlow\cite{wang2018}       & 3.30 & 12.95&  --  & 21.30&   --  & 5.23 &8.02  & 6.34 &9.08  \\
& OAFlow+ft\cite{wang2018}    &  --  & 3.55 & 4.2  & 8.88 & 31.2\%&(4.03)&7.95  &(5.95)&9.15  \\
& Back2Future-ft\cite{unflow_multi_occ}    &  --  &  --  &  --  & 6.59 &22.94\%&(3.89)&7.23  &(5.52)&8.81  \\
& DDFlow\cite{Pengpeng2019}                      &2.97  &8.27  &  --  &17.26 &   --  &3.83  &  --  & 4.85 &  --  \\
& DDFlow+ft\cite{Pengpeng2019}                   &  --  &2.35  &3.0   &5.72  &\textbf{14.29}\%&(2.92)&6.18  &\textbf{(3.98)}&7.40  \\
& EpiFlow\cite{Epipolar_flow_2019cvpr}&  --  &(2.51)&3.4   &(5.55)&16.95\%&(3.54)&7.00  &(4.99)&8.51  \\
& SelFlow\cite{Liu2019CVPR}           &  --  &1.97  &  --  &5.85  &   --  &(2.96)&  --  &(4.06)&  --  \\
& Ours                                     &\textbf{2.62}  & \textbf{1.78} & \textbf{2.1} & \textbf{4.57} & 15.20\%&(\textbf{2.82})&\textbf{5.79}  &4.13&\textbf{7.28}  \\
\bottomrule
\end{tabular}
}
\vspace{1mm}
\caption{Comparison with existing methods. Our method outperforms all the unsupervised optical flow methods on Flying Chairs and KITTI data sets. We report EPE (the lower the better) value as results on all the benchmarks except the test set of KITTI2015 where F1 (the lower the better) measurement is used by following the online evaluation. Missing entries `$-$' indicates that the results are not reported, and $(\cdot)$ indicates that the testing is conducted on the training data set. Note that SelFlow~\cite{Liu2019CVPR} is a multi-frame algorithm, we report its two-frame results for fair comparison. }\vspace{-5mm}
\label{comparision_sota}
\end{table}

\vspace{-2mm}
\subsection{Ablation Studies} \vspace{-2mm}
\paragraph{Boundary dilated warping.}
The first row and the fifth row in Table~\ref{ablation_study_component} show that the EPE value is significantly reduced by only replace the conventional warp function with our boundary dilated warp. In particular, KITTI data set is collected during driving with camera shakes, resulting in large number of pixels moving outside the image boundary. The proposed boundary dilated warp significantly improves the EPE from $6.12$ to $1.95$ in KITTI2012 and from $10.32$ to $5.00$ in KITTI2015, indicating the effectiveness of our boundary dilated warp. \vspace{-3mm}


\paragraph{Excluding photometric loss.}
Comparing row $1$ and row $2$, row $5$ and row $6$ in Table~\ref{ablation_study_component}, we can see that simply excluding object occlusion regions in the photometric loss can improve the performance on all the data sets. The reason is that the erroneous guidance of the photo loss is removed, making the learned optical flow more accurate in non-occluded regions. \vspace{-3mm}

\begin{figure}[t]
  \centering
  \includegraphics[width=0.99\textwidth]{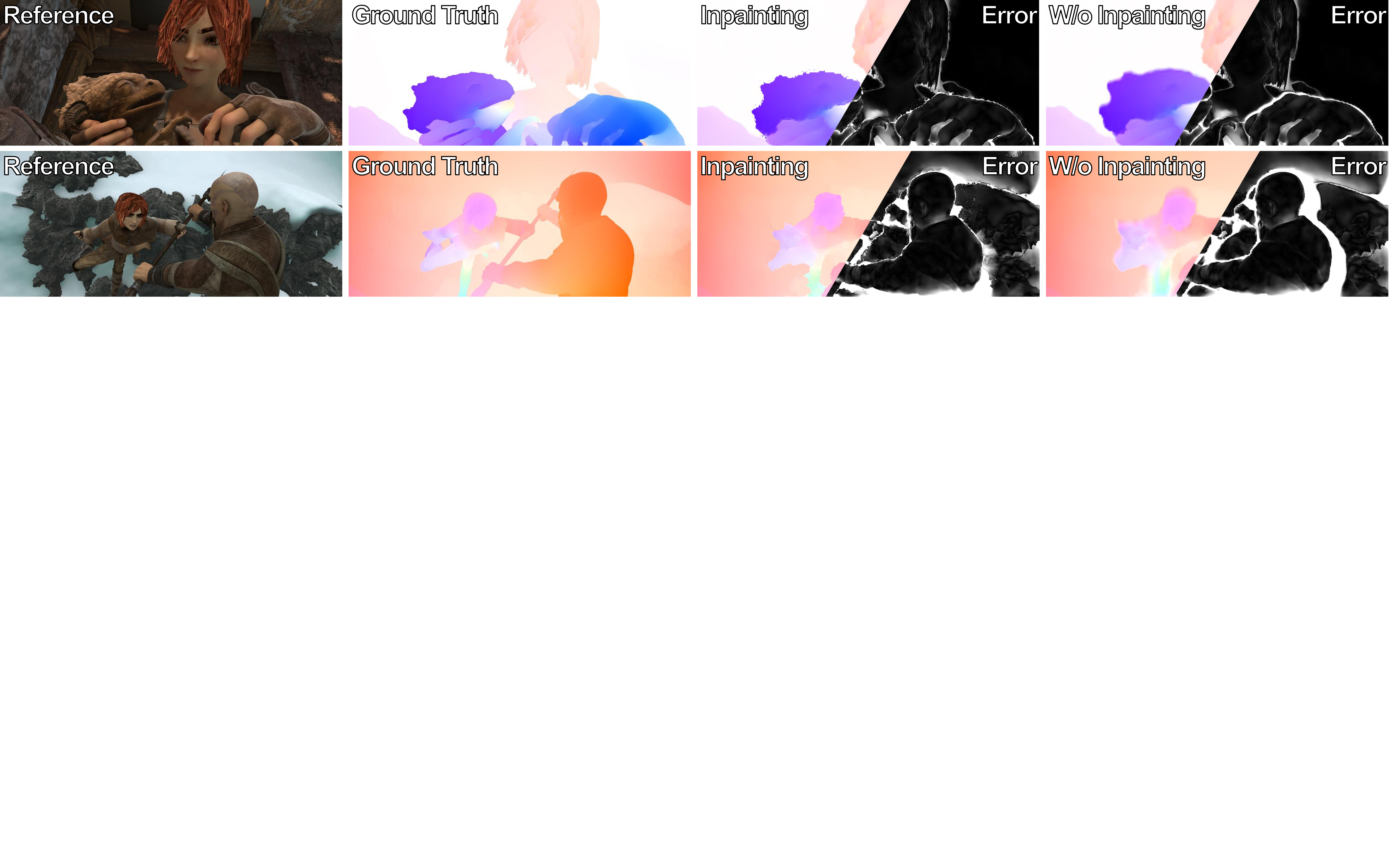}\\
  \vspace{-2mm}
  \caption{Comparison of whether using the occlusion inpainting. Results from Sintel Clean(top), and Sintel Final(bottom) are shown. }\label{fig:abl_result_on_sintal}\vspace{-2mm}
\end{figure}

\paragraph{Occlusion inpainting. }
Comparing row $1$ and row $3$, row $5$ and row $7$ in Table~\ref{ablation_study_component}, we can see that the performance can be improved by the occlusion inpainting method. Furthermore, comparing row $2$ and row $4$, row $6$ and row $8$ in Table~\ref{ablation_study_component}, we can see that excluding the photometric loss while using occlusion inpainting can further improve performance. This is because excluding photometric loss can only remove the incorrect guidance, and our occlusion inpainting method can introduce correct guidance for optical flow learning in occlusion regions. Some results are shown in Fig.~\ref{fig:abl_result_on_sintal}.
\vspace{-3mm}

\begin{table}[!t]
\centering
\resizebox*{1.0\linewidth}{!}{
\begin{tabular}{
>{\centering\arraybackslash}p{1.8cm}
>{\centering\arraybackslash}p{2.4cm}
>{\centering\arraybackslash}p{1.8cm}
>{\centering\arraybackslash}p{1.4cm}
>{\centering\arraybackslash}p{1.8cm}
>{\centering\arraybackslash}p{1.8cm}
>{\centering\arraybackslash}p{1.8cm}
>{\centering\arraybackslash}p{1.8cm}}
\toprule
boundary  & excluding  & occlusion   & Chairs& KITTI 2012 & KITTI 2015 & Sintel Clean & Sintel Final \\
dilated warp  & photometric loss  & inpainting  & test&  train & train &  train &  train \\
\midrule
           &           &           &   3.98    &   6.12    &   10.32   &    4.15   &   5.34    \\
           &\checkmark &           &   3.74    &   6.02    &    9.93   &    3.94   &   5.08    \\
           &           &\checkmark &   3.44    &   5.81    &   10.30   &    3.87   &   4.86    \\
           &\checkmark &\checkmark &   3.35    &   5.75    &    9.88   &    3.72   &   4.72    \\
\checkmark &           &           &   3.23    &   1.95    &   5.00    &    3.14   &   4.76    \\
\checkmark &\checkmark &           &   2.99    &   1.80    &   4.65    &    2.95   &   4.25    \\
\checkmark &           &\checkmark &   2.70    &   1.94    &   4.98    &    2.91   &   4.23    \\
\checkmark &\checkmark &\checkmark &   \textbf{2.62}    &   \textbf{1.78}    &   \textbf{4.57}    &    \textbf{2.82}   &   \textbf{4.13}    \\
\bottomrule
\end{tabular}}
\caption{Ablation study of components in the proposed method. }
\label{ablation_study_component}
\vspace{-6mm}

\end{table}

\section{Conclusion}\label{sec:conclu}\vspace{-3mm}
We have presented OccInpFlow, an occlusion-inpainting optical flow estimation framework for unsupervised learning. It works in a multi-scale manner and is composed of 3 main parts, including a feature extractor to extract the feature maps of the input frames in multiple scales, and a shared initial optical flow estimation block together with an optical flow refinement block working in each scale. Our framework enables the awareness the occlusion caused by internal objects and the limited image plane. We design a special boundary dilated warping combined with a light version of AppFlowNet to inpaint the occluded regions. We conduct extensive ablation studies to demonstrate the newly-introduced blocks can effectively improve the quality of the predicted optical flows. The results also show that our method significantly outperforms the state-of-the-art unsupervised approach for the optical flow estimation.
\vspace{-2mm}


\begin{thebibliography}{10}\itemsep=-1pt
	
	\bibitem{unsup_ICIP_2016}
	Aria Ahmadi and Ioannis Patras.
	\newblock Unsupervised convolutional neural networks for motion estimation.
	\newblock In {\em {Proc. ICIP}}, 2016.
	
	\bibitem{Alvarez2007}
	L. Alvarez, R. Deriche, T. Papadopoulo, and J. Sanchez.
	\newblock Symmetrical dense optical flow estimation with occlusions detection.
	\newblock {\em {International Journal of Computer Vision}}, 75(3):371--385,
	2007.
	
	\bibitem{Ayvaci2010}
	A. Ayvaci, M. Raptis, and S. Soatto.
	\newblock Occlusion detection and motion estimation with convex optimization.
	\newblock In {\em {Proc. NeurIPS}}, pages 100--108, 2010.
	
	\bibitem{cnn_patch_match2017}
	Christian Bailer, Kiran Varanasi, and Didier Stricker.
	\newblock Cnn-based patch matching for optical flow with thresholded hinge
	embedding loss.
	\newblock In {\em {Proc. CVPR}}, pages 2710--2719, 2017.
	
	\bibitem{bonneel2015blind}
	Nicolas Bonneel, James Tompkin, Kalyan Sunkavalli, Deqing Sun, Sylvain Paris,
	and Hanspeter Pfister.
	\newblock Blind video temporal consistency.
	\newblock {\em {ACM Trans. Graphics}}, 34(6):196:1--196:9, 2015.
	
	\bibitem{Thomas2004}
	Thomas Brox, Andres Bruhn, Nils Papenberg, and Joachim Weickert.
	\newblock High accuracy optical flow etimation based on a theory for warping.
	\newblock In {\em {Proc. ECCV}}, 2004.
	
	\bibitem{Thomas2015}
	Thomas Brox and Jitendra Malik.
	\newblock Large displacement optical flow: descriptor matching in variational
	motion estimation.
	\newblock {\em {IEEE Trans. on Pattern Analysis and Machine Intelligence}},
	33(3):500--513, 2011.
	
	\bibitem{Butler2012}
	Daniel Butler, Jonas Wulff, Garrett Stanley, and Michael Black.
	\newblock A naturalistic open source movie for optical flow evaluation.
	\newblock In {\em {Proc. ECCV}}, pages 611--625, 2012.
	
	\bibitem{Flownet_flyingchairs}
	Alexey Dosovitskiy, Philipp Fischer, Eddy Ilg, Philip Hausser, Caner Hazirbas,
	Vladimir Golkov, Patrick van~der Smagt, Daniel Cremers, and Thomas Brox.
	\newblock Flownet: Learning optical flow with convolutional networks.
	\newblock In {\em {Proc. ICCV}}, pages 2758--2766, 2015.
	
	\bibitem{KITTI_2012}
	A. Geiger, P. Lenz, and R. Urtasun.
	\newblock Are we ready for autonomous driving? the kitti vision benchmark
	suite.
	\newblock In {\em {Proc. CVPR}}, pages 3354--3361, 2012.
	
	\bibitem{Berthold1981}
	Berthold~KP Horn and Brian~G Schunck.
	\newblock Determining optical flow.
	\newblock {\em Artificial intelligence}, 17:1--3, 1981.
	
	\bibitem{LiteFlowNet}
	Tak-Wai Hui, Xiaoou Tang, and Chen~Change Loy.
	\newblock Liteflownet: A lightweight convolutional neural network for optical
	flow estimation.
	\newblock In {\em {Proc. CVPR}}, pages 8981--8989, 2018.
	
	\bibitem{Lightweight_tpami}
	Tak-Wai Hui, Xiaoou Tang, and Chen~Change Loy.
	\newblock A lightweight optical flow cnn - revisiting data fidelity and
	regularization.
	\newblock {\em {IEEE Trans. on Pattern Analysis and Machine Intelligence}},
	pages 1--1, 2020.
	
	\bibitem{Hur2017MirrorFlowES}
	Junhwa Hur and Stefan Roth.
	\newblock Mirrorflow: Exploiting symmetries in joint optical flow and occlusion
	estimation.
	\newblock In {\em {Proc. ICCV}}, pages 312--321, 2017.
	
	\bibitem{irrpwc}
	Junhwa Hur and Stefan Roth.
	\newblock Iterative residual refinement for joint optical flow and occlusion
	estimation.
	\newblock In {\em {Proc. CVPR}}, pages 5747--5756, 2019.
	
	\bibitem{FlowNet2}
	Eddy Ilg, Nikolaus Mayer, Tonmoy Saikia, Margret Keuper, Alexey Dosovitskiy,
	and Thomas Brox.
	\newblock Flownet 2.0: Evolution of optical flow estimation with deep networks.
	\newblock In {\em {Proc. CVPR}}, pages 1647--1655, 2017.
	
	\bibitem{unflow_multi_occ}
	Joel Janai, Fatma Güney, Anurag Ranjan, Michael Black, and Andreas Geiger.
	\newblock Unsupervised learning of multi-frame optical flow with occlusions.
	\newblock In {\em {Proc. ECCV}}, pages 713--731, 2018.
	
	\bibitem{Kroeger2016}
	Till Kroeger, Radu Timofte, Dengxin Dai, and Luc~Van Gool.
	\newblock Fast optical flow using dense inverse search.
	\newblock In {\em {Proc. ECCV}}, pages 471--488, 2016.
	
	\bibitem{semiflow_2017_nips}
	Wei-Sheng Lai, Jia-Bin Huang, and Ming-Hsuan Yang.
	\newblock Semi-supervised learning for optical flow with generative adversarial
	networks.
	\newblock In {\em {Proc. NeurIPS}}, pages 354--364, 2017.
	
	\bibitem{Liu2019UnsupervisedLO}
	Liang Liu, Guangyao Zhai, Wenlong Ye, and Yong Liu.
	\newblock Unsupervised learning of scene flow estimation fusing with local
	rigidity.
	\newblock In {\em IJCAI}, 2019.
	
	\bibitem{Pengpeng2019}
	Pengpeng Liu, Irwin King, Michael Lyu, and Jia Xu.
	\newblock Ddflow:learning optical flow with unlabeled data distillation.
	\newblock In {\em {Proc. AAAI}}, pages 8770--8777, 2019.
	
	\bibitem{Liu2019CVPR}
	Pengpeng Liu, Michael Lyu, Irwin King, and Jia Xu.
	\newblock Selflow:self-supervised learning of optical flow.
	\newblock In {\em {Proc. CVPR}}, pages 4571--4580, 2019.
	
	\bibitem{unflow_2018aaai}
	Simon Meister, Junhwa Hur, and Stefan Roth.
	\newblock Unflow: Unsupervised learning of optical flow with a bidirectional
	census loss, 2017.
	
	\bibitem{KITTI_2015}
	Moritz Menze and Andreas Geiger.
	\newblock Object scene flow for autonomous vehicles.
	\newblock In {\em {Proc. CVPR}}, pages 3061--3070, 2015.
	
	\bibitem{spynet2017}
	Anurag Ranjan and Michael~J. Black.
	\newblock Optical flow estimation using a spatial pyramid network.
	\newblock In {\em {Proc. CVPR}}, pages 2720--2729, 2017.
	
	\bibitem{Anurag2019}
	Anurag Ranjan, Varun Jampani, Lukas Balles, Kihwan Kim, Deqing Sun, Jonas
	Wulff, and Michael~J Black.
	\newblock Competitive collaboration: Joint unsupervised learning of depth,
	camera motion, optical flow and motion segmentation.
	\newblock In {\em {Proc. CVPR}}, pages 12240--12249, 2019.
	
	\bibitem{Ren_2019}
	Yurui Ren, Xiaoming Yu, Ruonan Zhang, Thomas~H Li, Shan Liu, and Ge Li.
	\newblock Structureflow: Image inpainting via structure-aware appearance flow.
	\newblock In {\em {Proc. ICCV}}, pages 181--190, 2019.
	
	\bibitem{Ren2017aaai}
	Zhe Ren, Junchi Yan, Bingbing Ni, Bin Liu, Xiaokang Yang, and Hongyuan Zha.
	\newblock Unsupervised deep learning for optical flow estimation.
	\newblock In {\em {Proc. AAAI}}, pages 1495--1501, 2017.
	
	\bibitem{EpicFlow_2015}
	Jerome Revaud, Philippe Weinzaepfel, Zaid Harchaoui, and Cordelia Schmid.
	\newblock Epicflow: Edge-preserving interpolation of correspondences for
	optical flow.
	\newblock In {\em {Proc. CVPR}}, pages 1164--1172, 2015.
	
	\bibitem{Sun2016}
	Laura Sevilla-Lara, Deqing Sun, Varun Jampani, and Michael~J Black.
	\newblock Optical flow with semantic segmentation and localized layers.
	\newblock In {\em {Proc. CVPR}}, pages 3889--3898, 2016.
	
	\bibitem{Simonyan2014}
	K. Simonyan and A. Zisserman.
	\newblock Two-stream convolutional networks for action recognition in videos.
	\newblock In {\em {Proc. NeurIPS}}, 2014.
	
	\bibitem{Sun2014cvpr}
	D. Sun, C. Liu, and H. Pfister.
	\newblock Local layering for joint motion estimation and occlusion detection.
	\newblock In {\em {Proc. CVPR}}, pages 1098--1105, 2014.
	
	\bibitem{Sun2014}
	D Sun, S Roth, and MJ Black.
	\newblock A quantitative analysis of current practices in optical flow
	estimation and the principles behind them.
	\newblock {\em {International Journal of Computer Vision}}, 106:115--137, 2014.
	
	\bibitem{Sun2010}
	Deqing Sun, Stefan Roth, and Michael~J Black.
	\newblock Secrets of optical flow estimation and their principles.
	\newblock In {\em {Proc. CVPR}}, pages 2432--2439, 2010.
	
	\bibitem{Sun2010nips}
	D. Sun, E.~B. Sudderth, and M.~J. Black.
	\newblock Layered image motion with explicit occlusions, temporal consistency
	and depth ordering.
	\newblock In {\em {Proc. NeurIPS}}, pages 2226--2234, 2010.
	
	\bibitem{pwc_net}
	Deqing Sun, Xiaodong Yang, Ming-Yu Liu, and Jan Kautz.
	\newblock Pwc-net: Cnns for optical flow using pyramid, warping, and cost
	volume.
	\newblock In {\em {Proc. CVPR}}, pages 8934--8943, 2018.
	
	\bibitem{pwc_net_tpami}
	Deqing Sun, Xiaodong Yang, Ming-Yu Liu, and Jan Kautz.
	\newblock Models matter, so does training: An empirical study of cnns for
	optical flow estimation.
	\newblock {\em {IEEE Trans. on Pattern Analysis and Machine Intelligence}},
	42(6):1408--1423, 2020.
	
	\bibitem{Wang2019UnOSUU}
	Yang Wang, Peng Wang, Zhenheng Yang, Chenxu Luo, Yezhou Yang, and Wei Xu.
	\newblock Unos: Unified unsupervised optical-flow and stereo-depth estimation
	by watching videos.
	\newblock In {\em {Proc. CVPR}}, pages 8063--8073, 2019.
	
	\bibitem{wang2018}
	Yang Wang, Yi Yang, Zhenheng Yang, Liang Zhao, Peng Wang, and Wei Xu.
	\newblock Occlusion aware unsupervised learning of optical flow.
	\newblock In {\em {Proc. CVPR}}, pages 4884--4893, 2018.
	
	\bibitem{CostVolume_cvpr2017}
	Jia Xu, Rene Ranftl, and Vladlen Koltun.
	\newblock Accurate optical flow via direct cost volume processing.
	\newblock In {\em {Proc. CVPR}}, pages 5807--5815, 2017.
	
	\bibitem{VolumetricCN2019}
	Gengshan Yang and Deva Ramanan.
	\newblock Volumetric correspondence networks for optical flow.
	\newblock In {\em {Proc. NeurIPS}}, pages 794--805, 2019.
	
	\bibitem{unsup_CPN}
	Yanchao Yang and Stefano Soatto.
	\newblock Conditional prior networks for optical flow.
	\newblock In {\em {Proc. ECCV}}, pages 282--298, 2018.
	
	\bibitem{Yin2018GeoNetUL}
	Zhichao Yin and Jianping Shi.
	\newblock Geonet: Unsupervised learning of dense depth, optical flow and camera
	pose.
	\newblock In {\em {Proc. CVPR}}, pages 1983--1992, 2018.
	
	\bibitem{Jason2016}
	Jason Yu, Adam Harley, and Konstantions Derpanis.
	\newblock Back to basics:unsupervised learning of optical flow via brightness
	constancy and motion smoothness.
	\newblock In {\em Proc. ECCV Workshops}, pages 3--10, 2016.
	
	\bibitem{zhao2020maskflownet}
	Shengyu Zhao, Yilun Sheng, Yue Dong, Eric I-Chao Chang, and Yan Xu.
	\newblock Maskflownet: Asymmetric feature matching with learnable occlusion
	mask.
	\newblock In {\em {Proc. CVPR}}, 2020.
	
	\bibitem{Epipolar_flow_2019cvpr}
	Yiran Zhong, Pan Ji, Jianyuan Wang, Yuchao Dai, and Hongdong Li.
	\newblock Unsupervised deep epipolar flow for stationary or dynamic scenes.
	\newblock In {\em {Proc. CVPR}}, pages 12095--12104, 2019.
	
	\bibitem{first_appearance_flow_for_view_synthesis}
	Tinghui Zhou, Shubham Tulsiani, Weilun Sun, Jitendra Malik, and Alexei~A.
	Efros.
	\newblock View synthesis by appearance flow.
	\newblock In {\em {Proc. ECCV}}, pages 286--301, 2016.
	
	\bibitem{dfnet_zou_2018}
	Yuliang Zou, Zelun Luo, and Jia-Bin Huang.
	\newblock Df-net: Unsupervised joint learning of depth and flow using
	cross-task consistency.
	\newblock In {\em {Proc. ECCV}}, pages 38--55, 2018.
	
\end{thebibliography}

\newpage
\part*{Appendix}
\begin{appendices}
\begin{figure}[ht]
  \centering
  \subfigure[Framework of AppFlowNet-Lite.]{
  \includegraphics[width=0.60\linewidth]{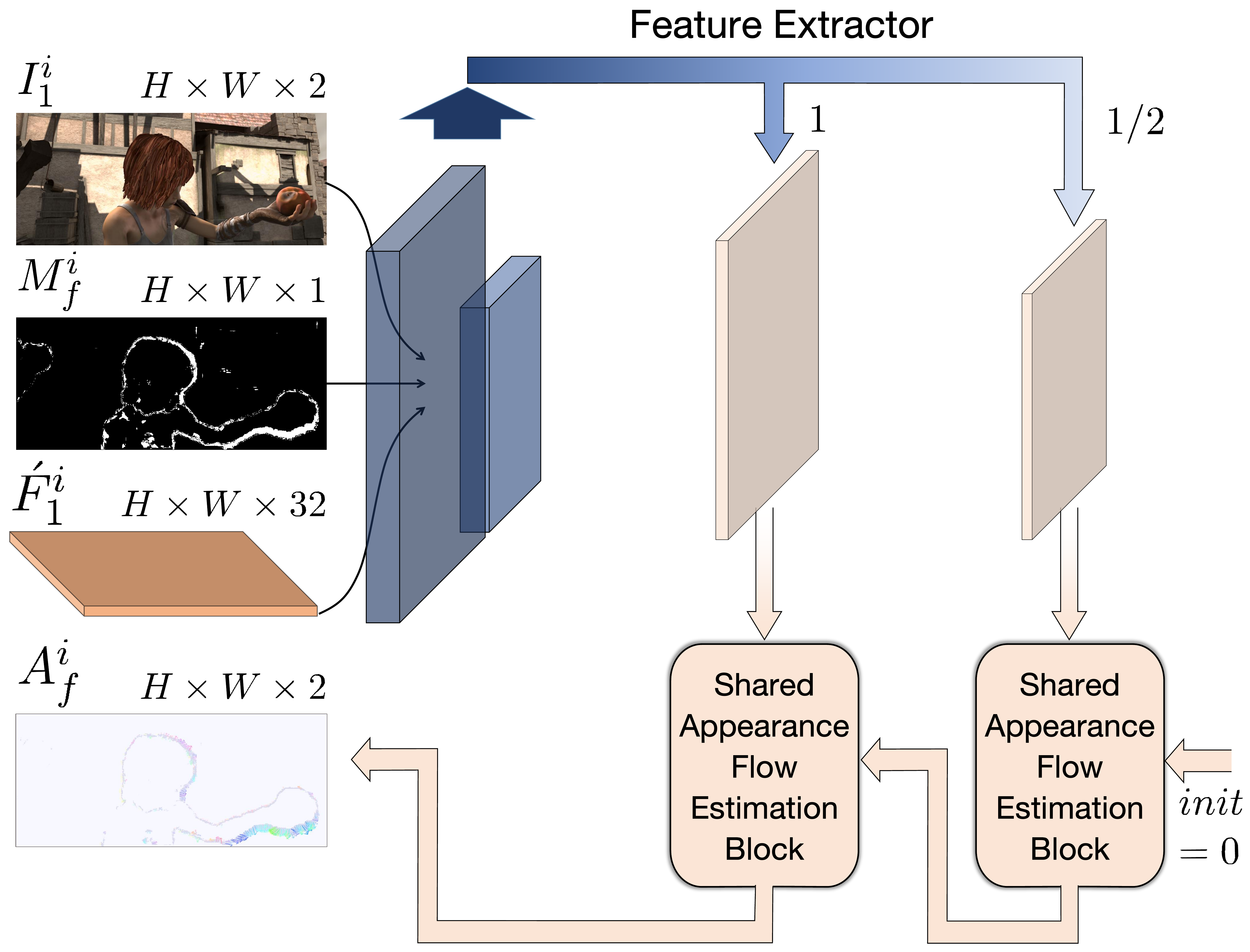}\label{subfig:appflownet_lite_frame}
  }
\hfill
  \subfigure[Appearance Flow Esitmation Block.]{
  \includegraphics[width=0.34\linewidth]{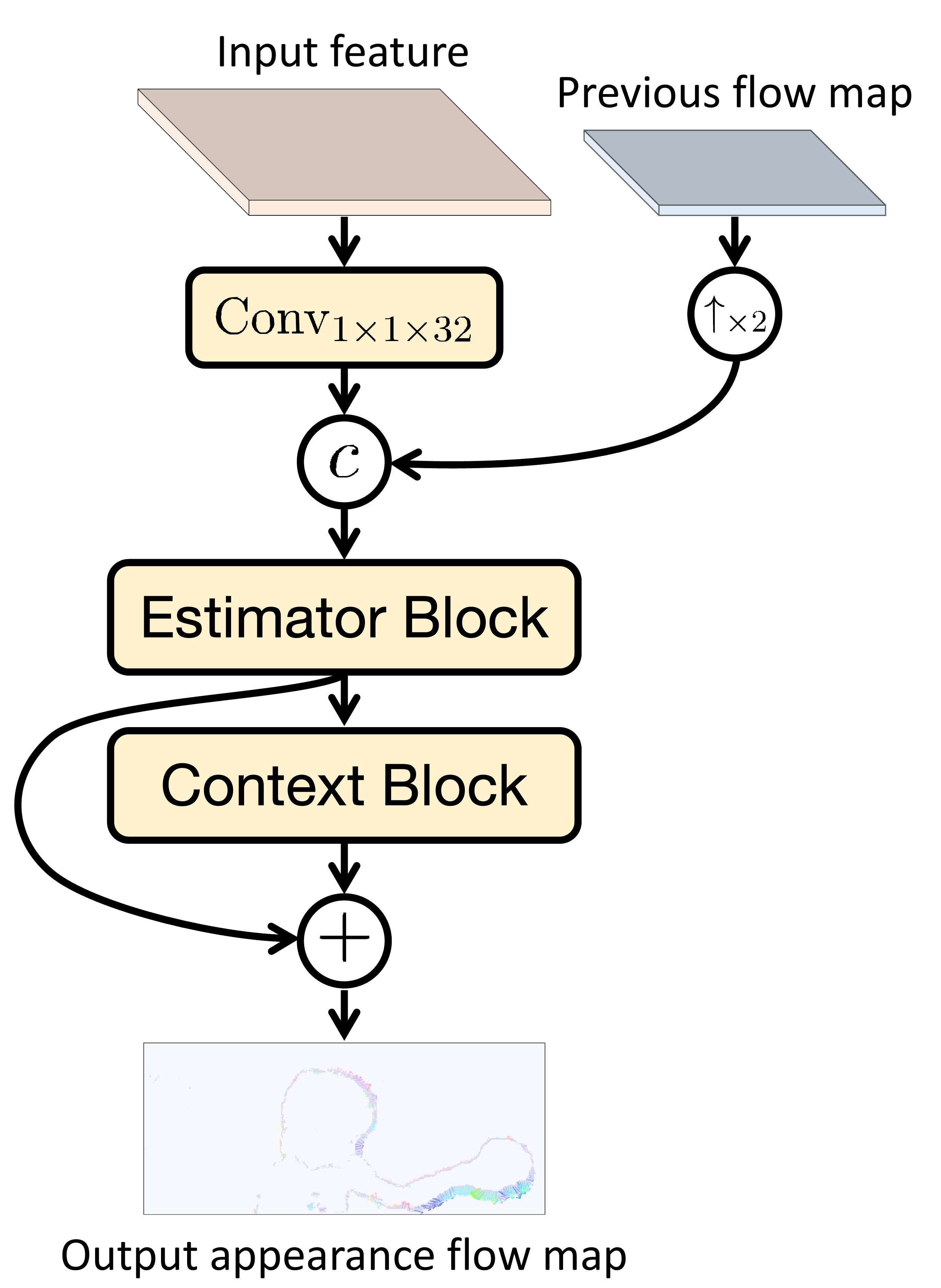}\label{subfig:appflownet_lite_decoder}
  }
  \vspace{-3mm}
  \caption{The framework of the AppFlowNet-Lite (a) and the appearance flow estimation block (b) embedded in (a). }\label{fig:appflownet_lite_structure}\vspace{-3mm}\label{fig:appflownet_lite}
\end{figure}

\section{Structure of AppFlowNet-Lite}\label{sec:appflownet_lite}\vspace{-2mm}
Our AppFlowNet-Lite in shown in Fig.~\ref{subfig:appflownet_lite_frame}. In the entire network, we first use several conv layers to extract two feature maps in $1$ and $1/2$ scale. Then, a shared decoder is utilized to produce the appearance flow. The detailed structure of the shared decoder is illustrated in Fig.~\ref{subfig:appflownet_lite_decoder}, where we first use a $1\times1$ conv layer to modify the number of the input feature channels into an uniform one, and then concatenate the unified feature with the previous flow map followed by feeding it into an estimator block to produce the appearance flow. Note that the initial flow map for decoder in $1/2$ scale is set as zeros. Also, to learn residual information, a context block is connected following the Estimator Block to better refine the appearance flow. As for the detailed configuration for the Estimator and Context Blocks, we simply use $5$ dense connected conv layers in the former one, and use $5$ dilated conv layers of kernel sizes of $\{1,2,4,8,16\}$ and two conv layers in the latter one, to produce the residual information.


\section{Boundary Dilated Warping}\label{sec:boundary_dilated_warping_example}
\begin{figure}[ht]
  \centering
  \includegraphics[width=0.9\linewidth]{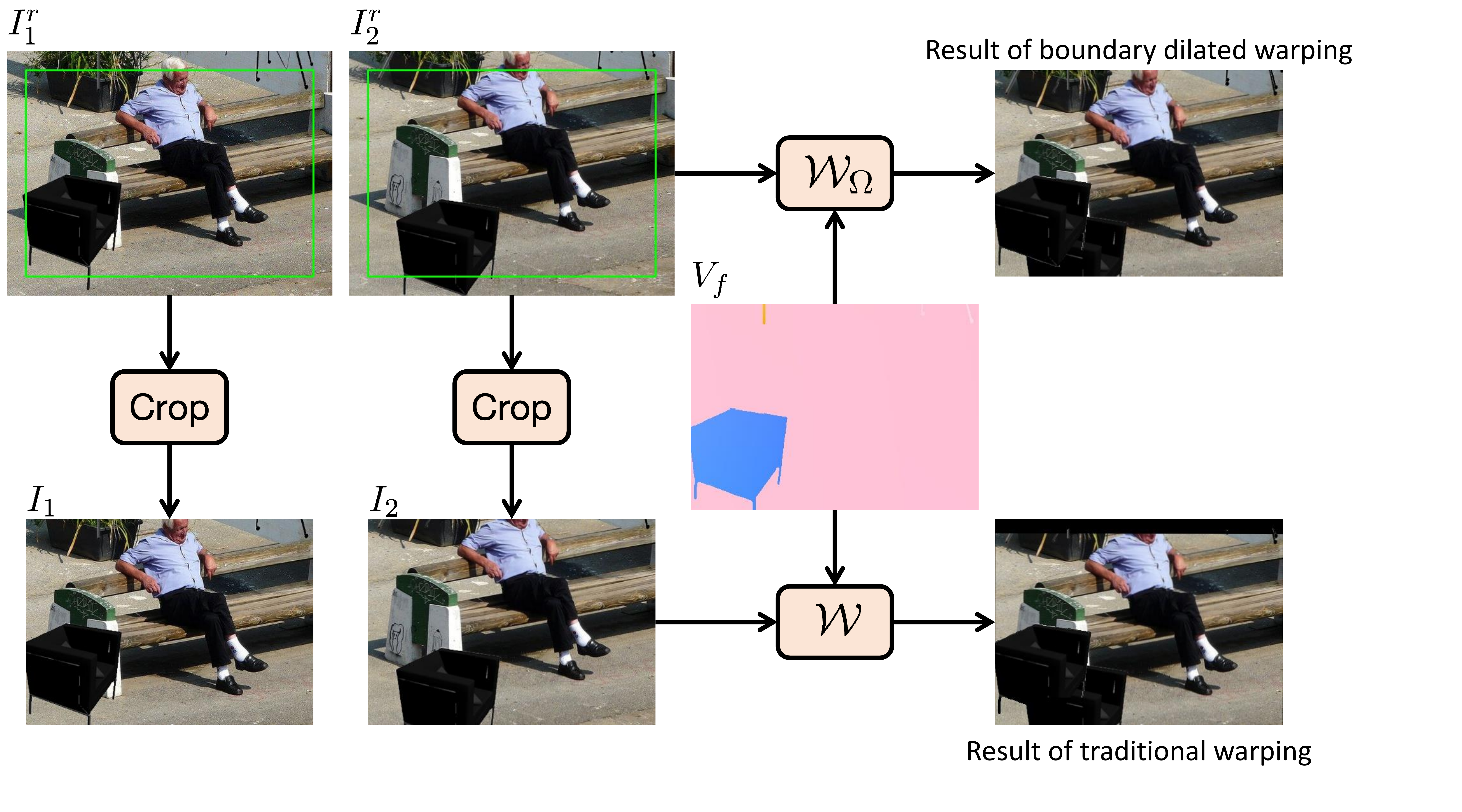}\\
  \caption{An example of the boundary dilated warping v.s. the traditional warping. }\label{fig:boundary_warp_example}
\end{figure}

\vspace{-2mm}
In order to further illustrate the proposed boundary dilated warp, we show an example in Fig~\ref{fig:boundary_warp_example}. The original reference image $I_1^r$ and target image $I_2^r$ are cropped into the training image pair $I_1$ and $I_2$. Our goal is to warp the target image $I_2$ based on the optical flow $V_f$. The traditional warping method $\mathcal{W}$ directly warp pixels of $I_2$ by $V_f$, ignoring motions toward the outside of the image plane. Thus the upper region of the warped image is filled with zeros, i.e. black pixels. To eliminate these shadows, we designed the boundary dilated warping method that warp pixels of $I_2^r$ based on $V_f$, and pixels moving outside the image plane of $I_2$ can be mapped back and used to fill in the missing regions.


\begin{figure}[ht]
  \centering
  \includegraphics[width=0.95\linewidth]{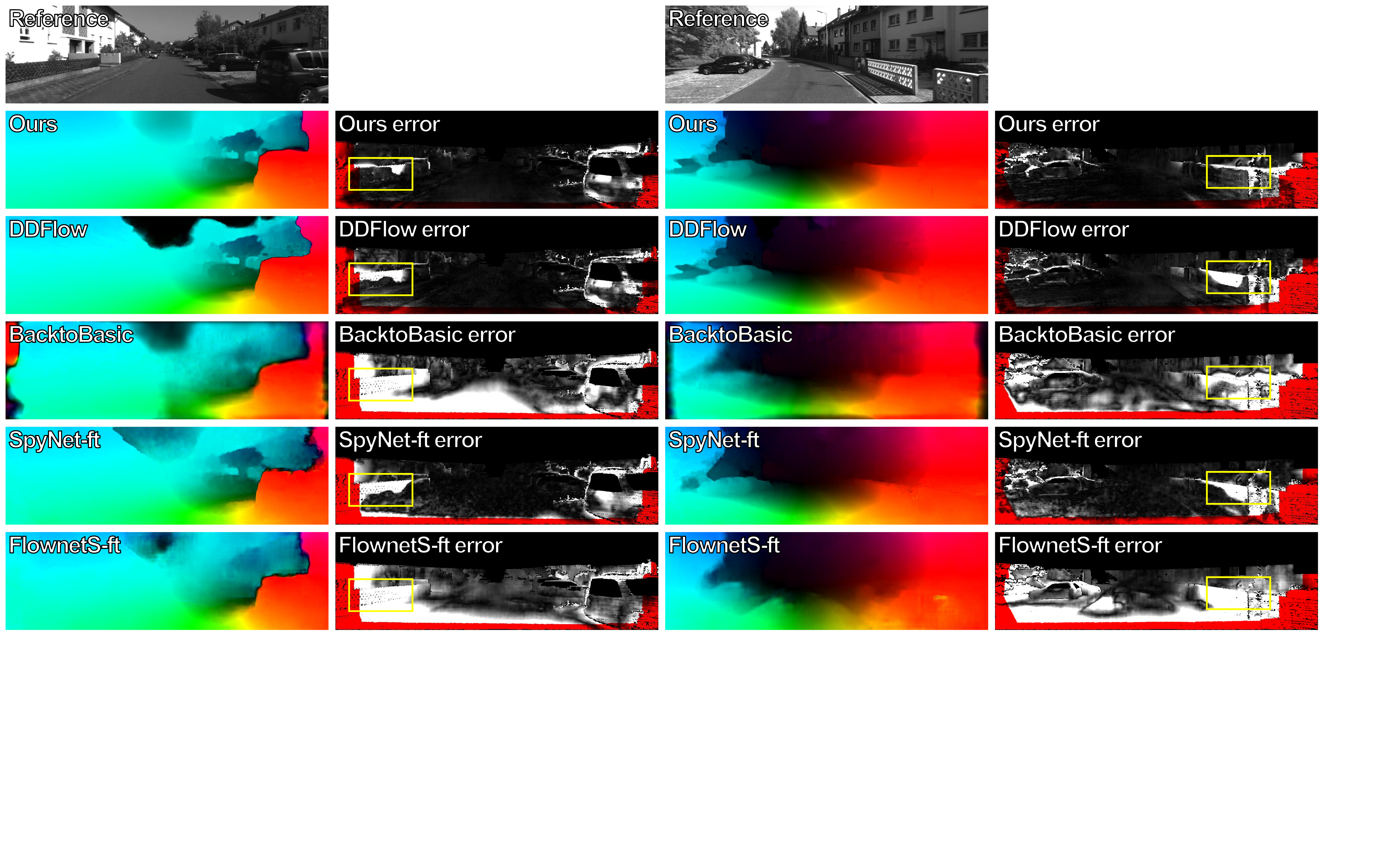}\\
  \caption{Comparison on KITTI 2012 benchmark. Results of the unsupervised methods DDFlow \cite{Pengpeng2019} and BacktoBasic \cite{Jason2016} and the fully supervised methods SpyNet \cite{spynet2017} and FlowNetS \cite{Flownet_flyingchairs} are shown. For better comparison, error maps are also visualized in the even columns, where areas with larger errors are brighter and the occluded regions are highlighted in red.  }\label{fig:qualitative_results_pk_kitti}
\end{figure}

%

\begin{figure}[ht]
	\centering
	\subfigure[MPI-Sintel Clean benchmark.]{
		\includegraphics[width=0.98\linewidth]{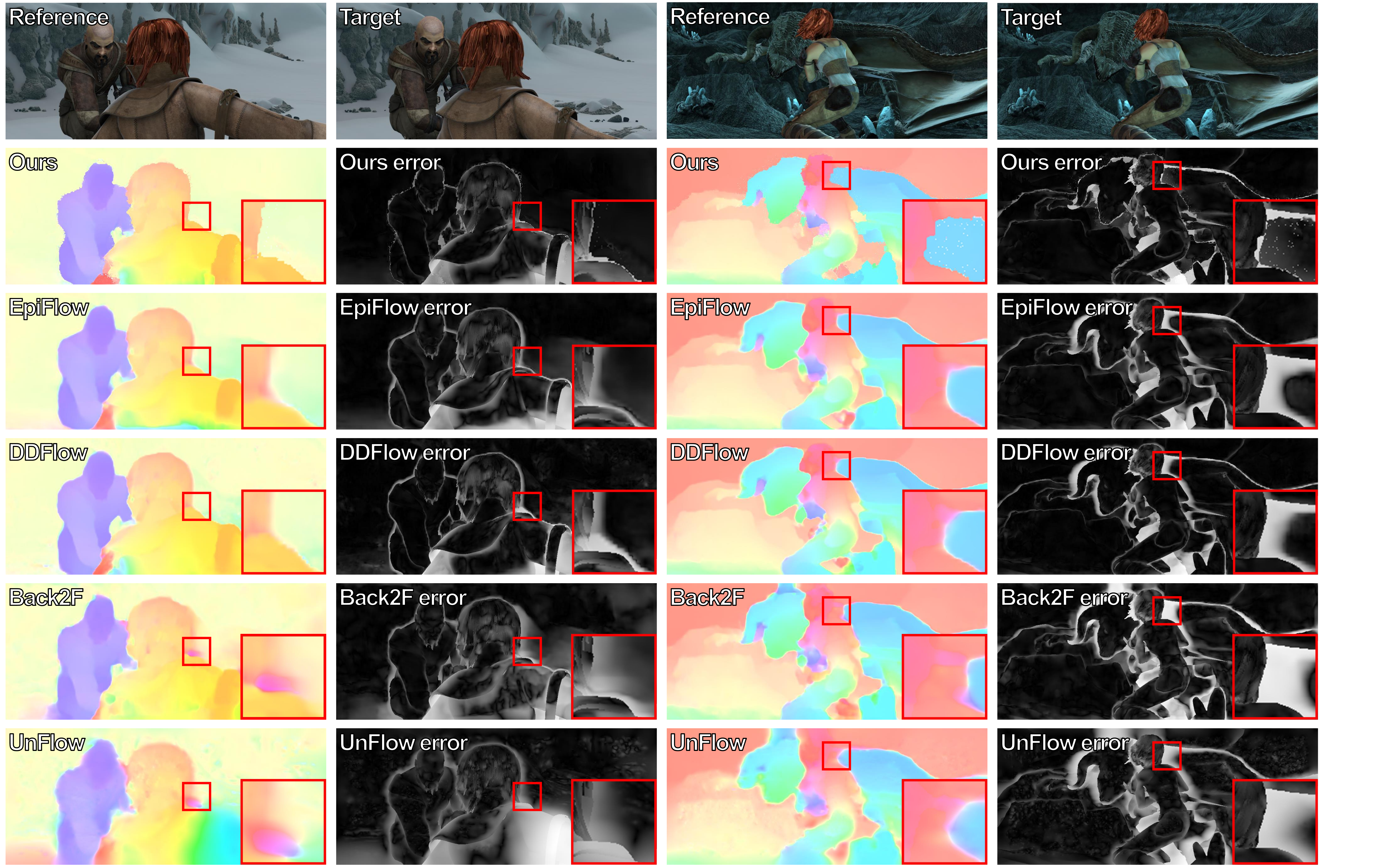}\label{fig:qualitative_results_pk_clean}
	}
	\subfigure[KITTI 2012 train set.]{
		\includegraphics[width=0.98\linewidth]{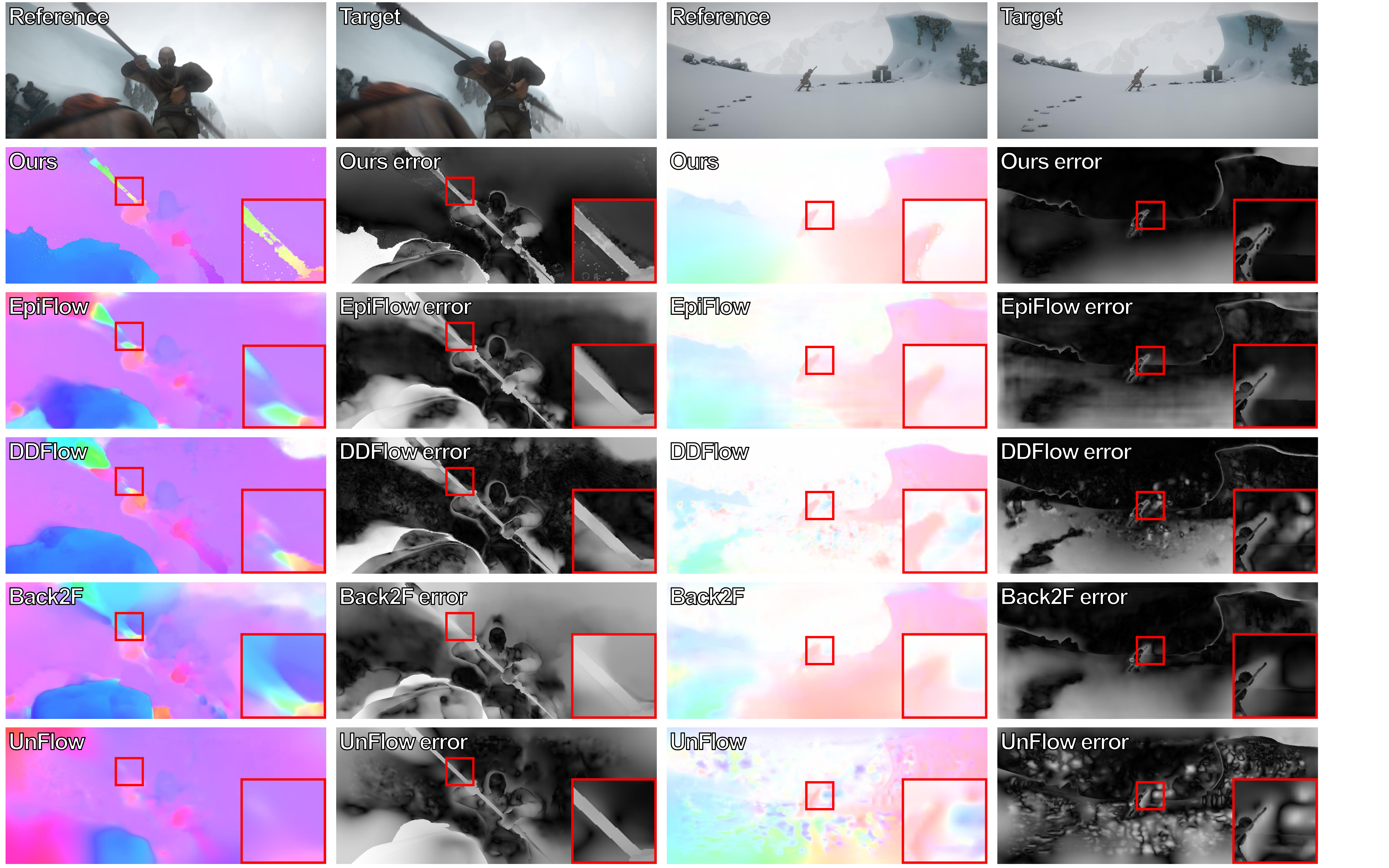}\label{fig:qualitative_results_pk_final}
	}
	\vspace{-3mm}
	\caption{Comparison on MPI-Sintel Clean benchmark(a) and MPI-Sintel Final benchmark(b), where results of more unsupervised methods EpiFlow \cite{Epipolar_flow_2019cvpr}, DDFlow \cite{Pengpeng2019}, Back2future \cite{unflow_multi_occ}, UnFlow \cite{unflow_2018aaai} are shown. The error maps of the predictions are visualized in the even columns. For better visualization, the zoom in views of some local patches are depicted in the right bottom corner of each sample, highlighted in red box.}\vspace{-3mm}
\end{figure}

\section{Qualitive Results}\label{sec:more_pk_results}

\subsection{KITTI 2012 Benchmark}\vspace{-2mm}
As shown in Fig.~\ref{fig:qualitative_results_pk_kitti}, we compare our method with several unsupervised methods such as DDFlow~\cite{Pengpeng2019}, BacktoBasic~\cite{Jason2016} and fully supervised ones such as SpyNet~\cite{spynet2017} and FlowNetS~\cite{Flownet_flyingchairs}. Our method produces smaller error than other unsupervised ones as emphasized in the yellow boxes. It is also worth noting that, our method even outperforms the two fully supervised ones.

\subsection{MPI-Sintel Benchmark}\vspace{-2mm}
We further compare our method with more unsupervised approaches including EpiFlow~\cite{Epipolar_flow_2019cvpr}, DDFlow~\cite{Pengpeng2019}, Back2future~\cite{unflow_multi_occ} and UnFlow~\cite{unflow_2018aaai}. As seen from Fig.~\ref{fig:qualitative_results_pk_clean} and~\ref{fig:qualitative_results_pk_final}, our method produces sharper predictions in object edges and less noise in other regions.

\section{Ablation Experiments}\label{sec:more_results}\vspace{-2mm}

\subsection{Occlusion Inpainting}
\begin{figure}[ht]
  \centering
  \subfigure[Flying Chairs test set.]{
  \includegraphics[width=0.98\linewidth]{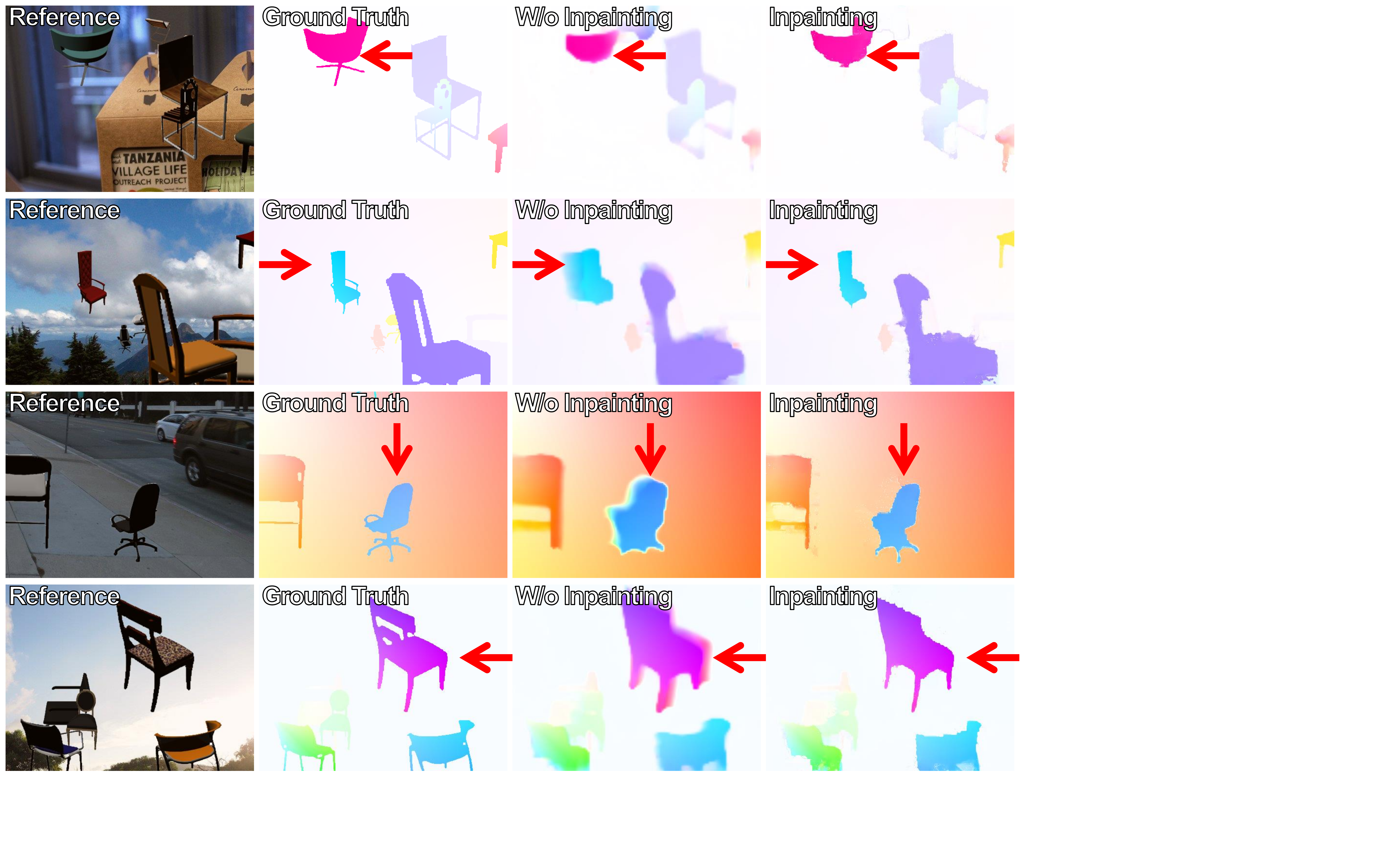}\label{fig:abl_result_if_appflow_flying_chairs}
  }
  \subfigure[KITTI 2012 train set.]{
  \includegraphics[width=0.98\linewidth]{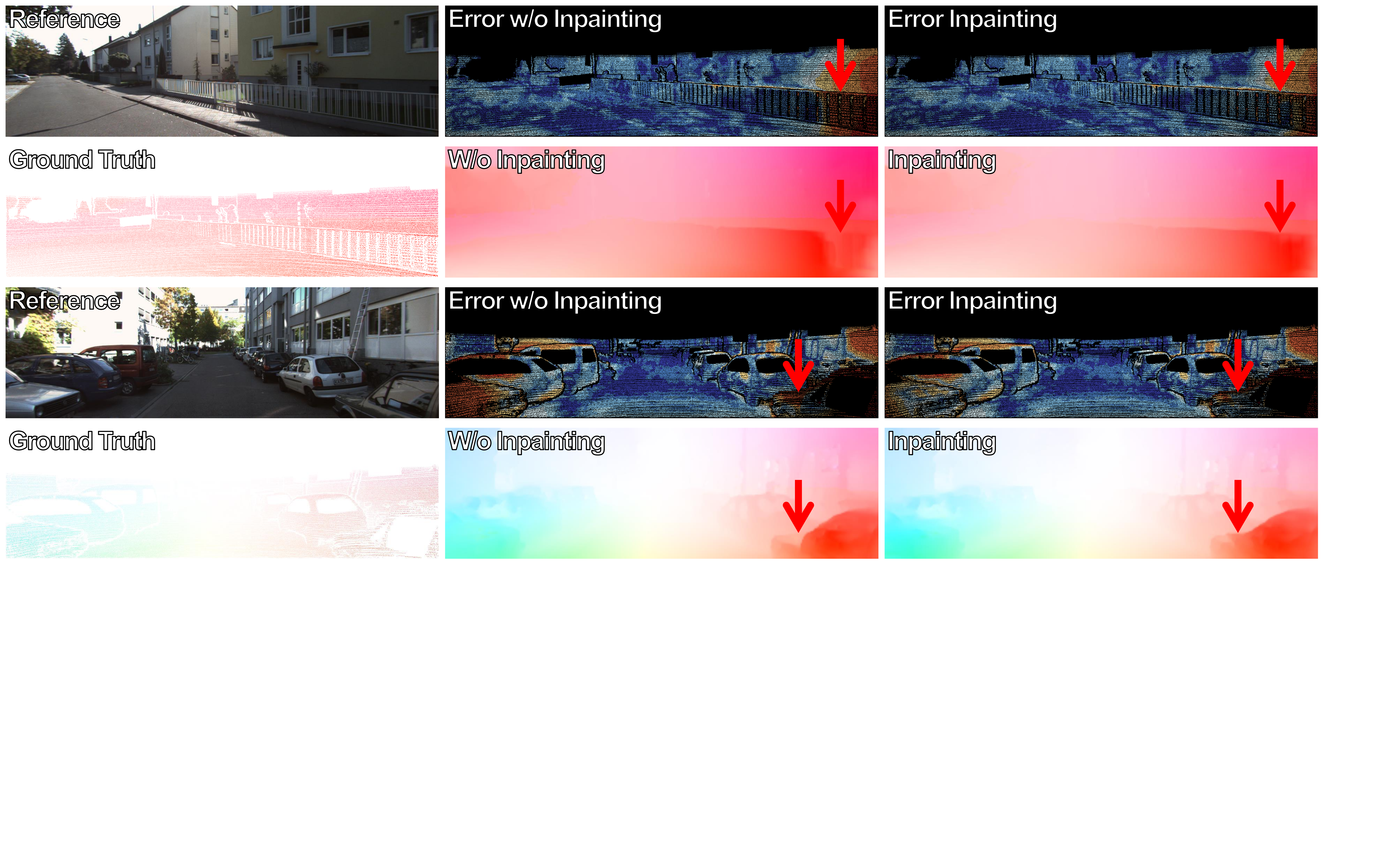}\label{fig:abl_result_if_appflow_kitti}
  }
  \vspace{-3mm}
  \caption{Comparison of with v.s. without the occlusion inpainting on Flying Chairs (a) and KITTI 2012 (b) data sets. The error maps are provided for better view where correct and incorrect predictions are dipicted in blue and red respectively. The obvious difference is highlighted by the red arrows. }\label{fig:abl_result_if_appflow_kitti_flying}\vspace{-3mm}
\end{figure}

\begin{figure}[ht]
  \centering
  \subfigure[Sintel Clean train set.]{
  \includegraphics[width=0.98\linewidth]{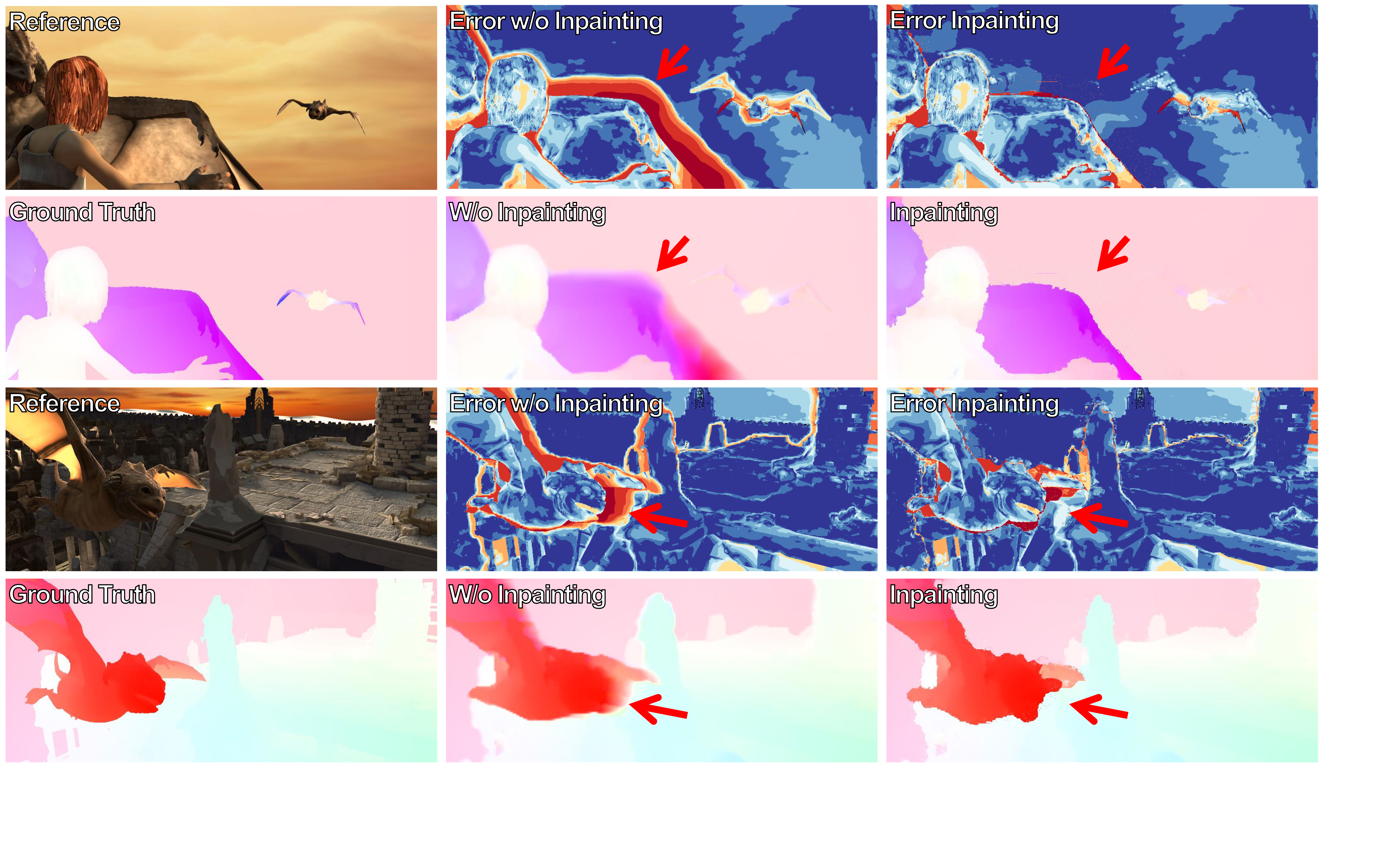}\label{fig:abl_result_if_appflow_Sintel_clean}
  }
  \subfigure[Sintel Final train set.]{
  \includegraphics[width=0.98\linewidth]{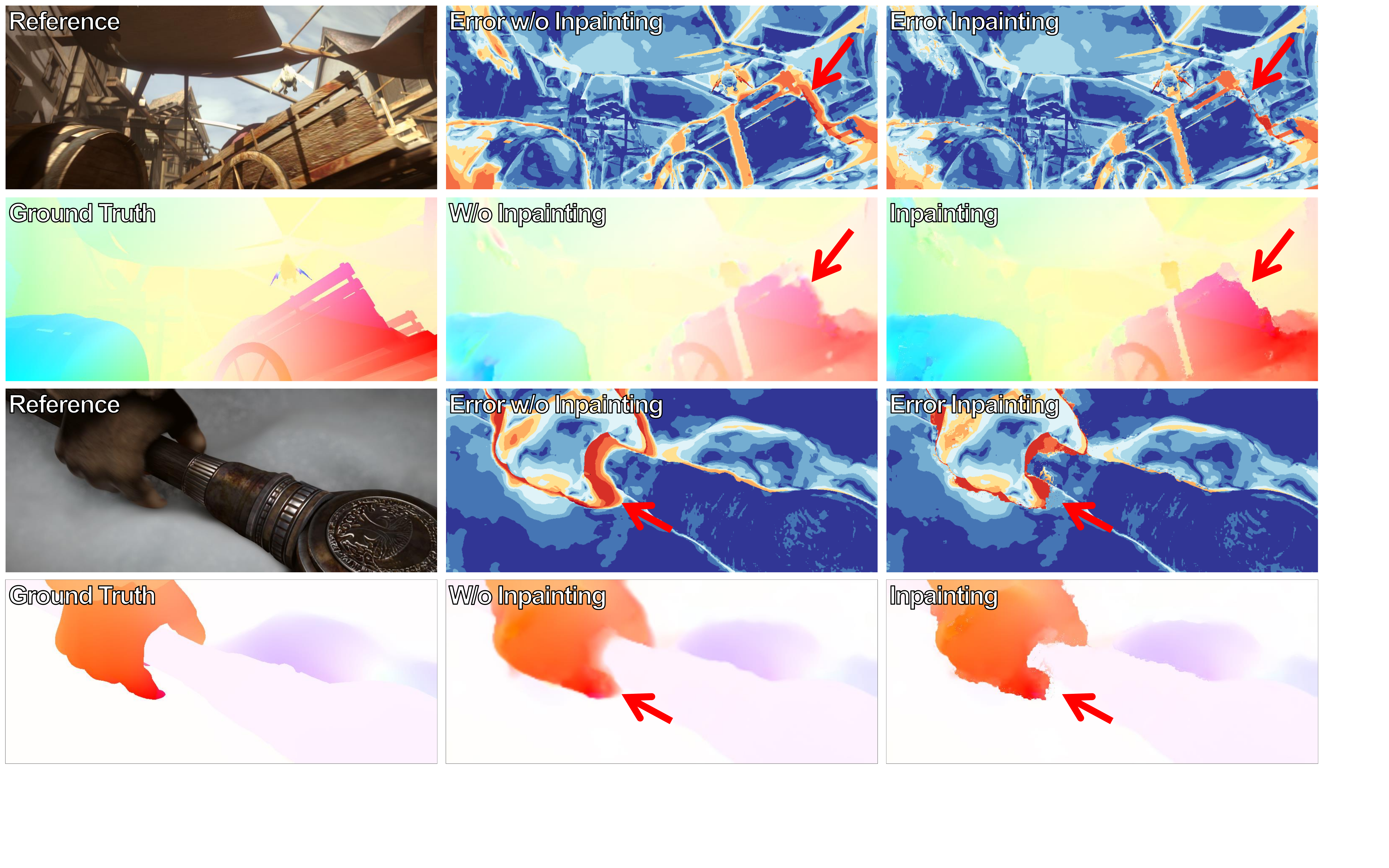}\label{fig:abl_result_if_appflow_Sintel_final}
  }
  \vspace{-3mm}
  \caption{Comparison of with v.s. without the occlusion inpainting on MPI-Sintel Clean (a) and Final (b) data sets. }\label{fig:abl_result_if_appflow}\vspace{-3mm}
\end{figure}

\vspace{-2mm}
We show results of enabling v.s. disabling our occlusion inpainting method on multiple datasets such as Flying Chairs (Fig.~\ref{fig:abl_result_if_appflow_flying_chairs}), KITTI 2012 (Fig.~\ref{fig:abl_result_if_appflow_kitti}), Sintel Clean (Fig.~\ref{fig:abl_result_if_appflow_Sintel_clean}) and Sintel Final (Fig.~\ref{fig:abl_result_if_appflow_Sintel_final}). As seen, optical flow predictions by our occlusion inpainting method are significantly sharper in object edges and more accurate in the occluded regions.

\subsection{Boundary Dilated Warping}
\begin{figure}[ht]
  \centering
  \subfigure[Flying Chairs test set.]{
  \includegraphics[width=0.98\linewidth]{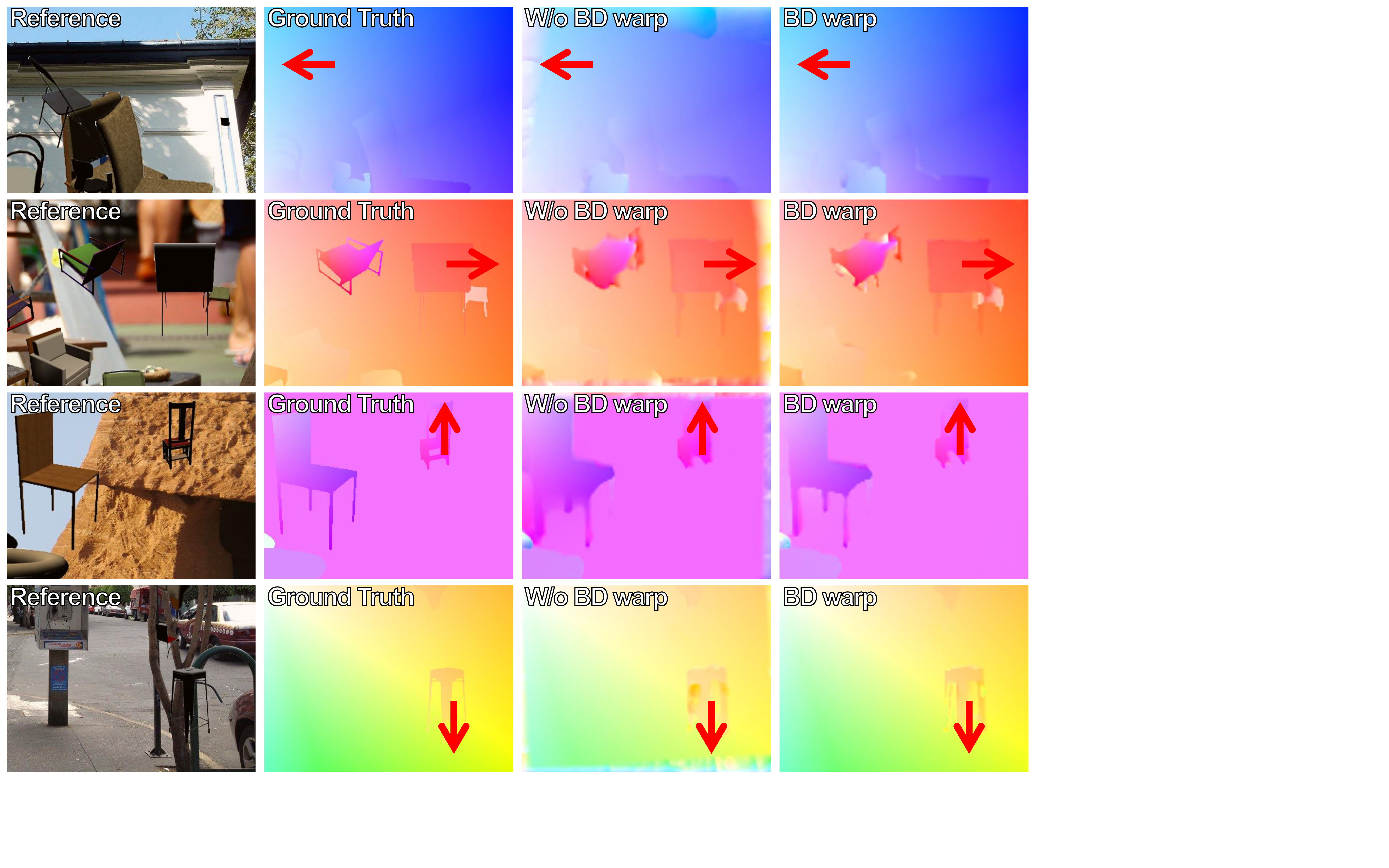}\label{fig:abl_result_if_bdwarp_flying_chairs}
  }
  \subfigure[KITTI 2012 train set.]{
  \includegraphics[width=0.98\linewidth]{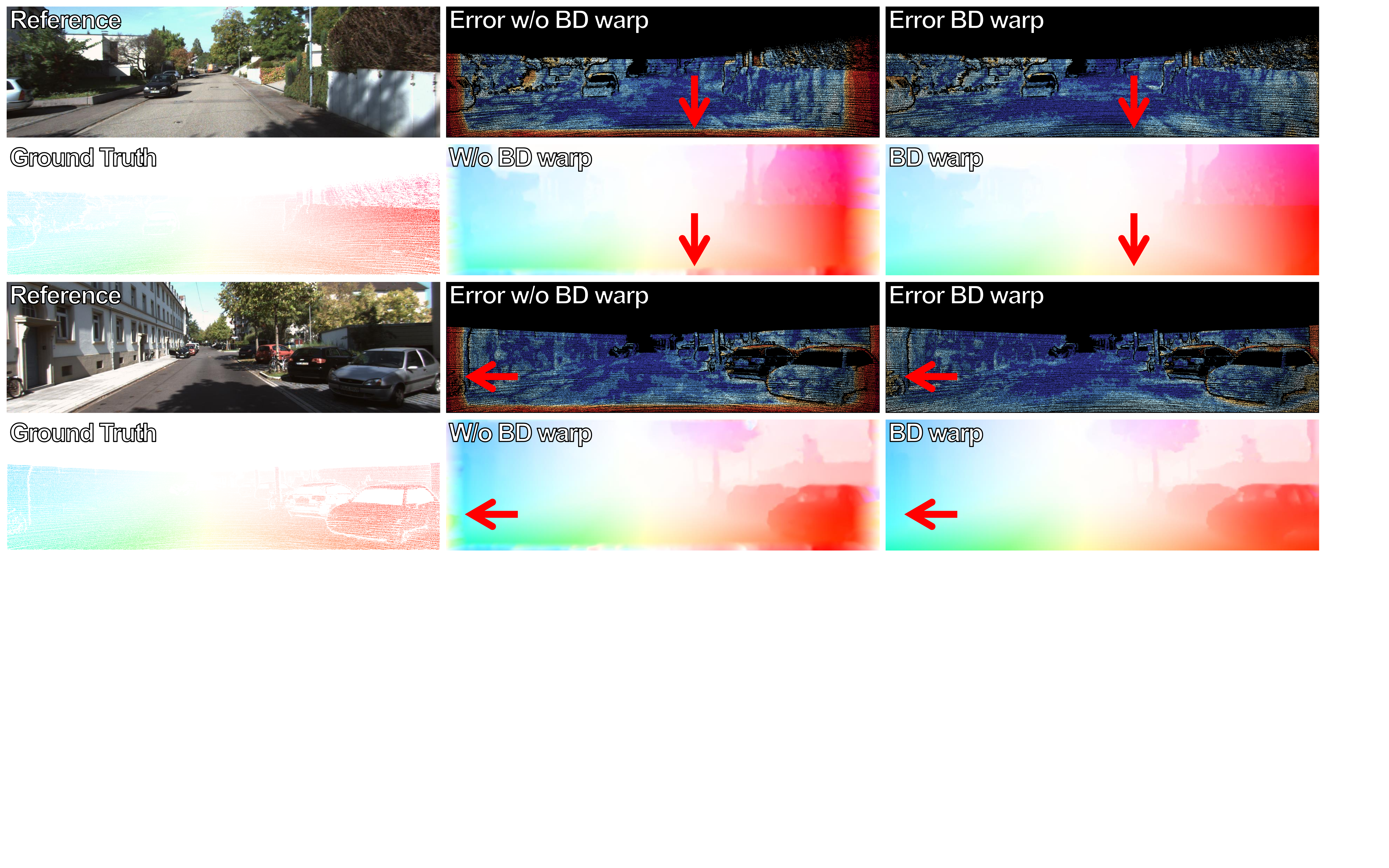}\label{fig:abl_result_if_bdwarp_kitti2012}
  }
  \vspace{-3mm}
    \caption{Comparison of with v.s. without the boundary dilated warping on Flying Chairs (a) and KITTI 2012 (b) data sets. }\label{fig:abl_result_if_bdwarp}\vspace{-3mm}
\end{figure}

\begin{figure}[ht]
  \centering
  \subfigure[Sintel Clean train set. ]{
  \includegraphics[width=0.98\linewidth]{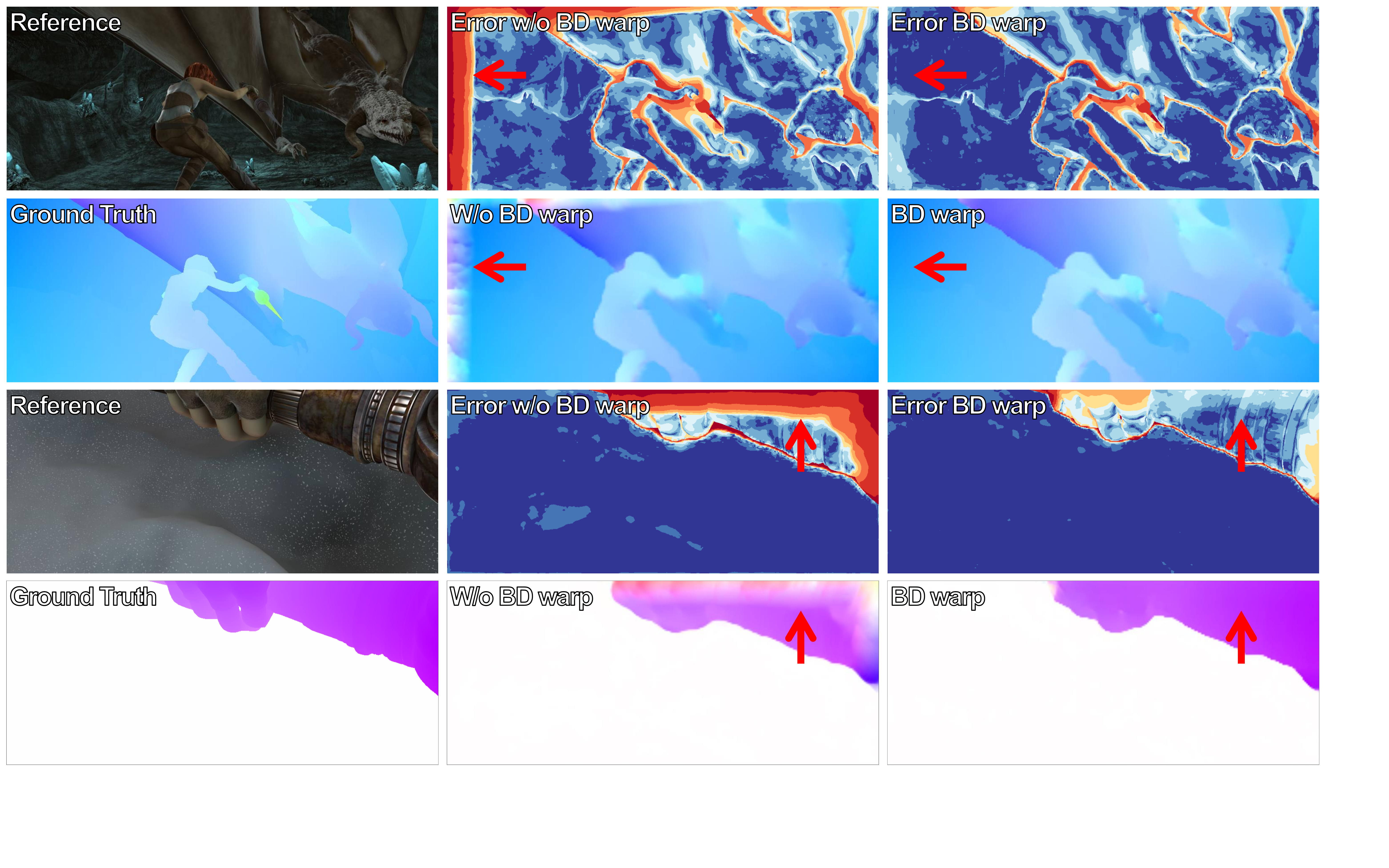}\label{fig:abl_result_if_bdwarp_Sintel_clean}
  }
  \subfigure[Sintel Final train set.]{
  \includegraphics[width=0.98\linewidth]{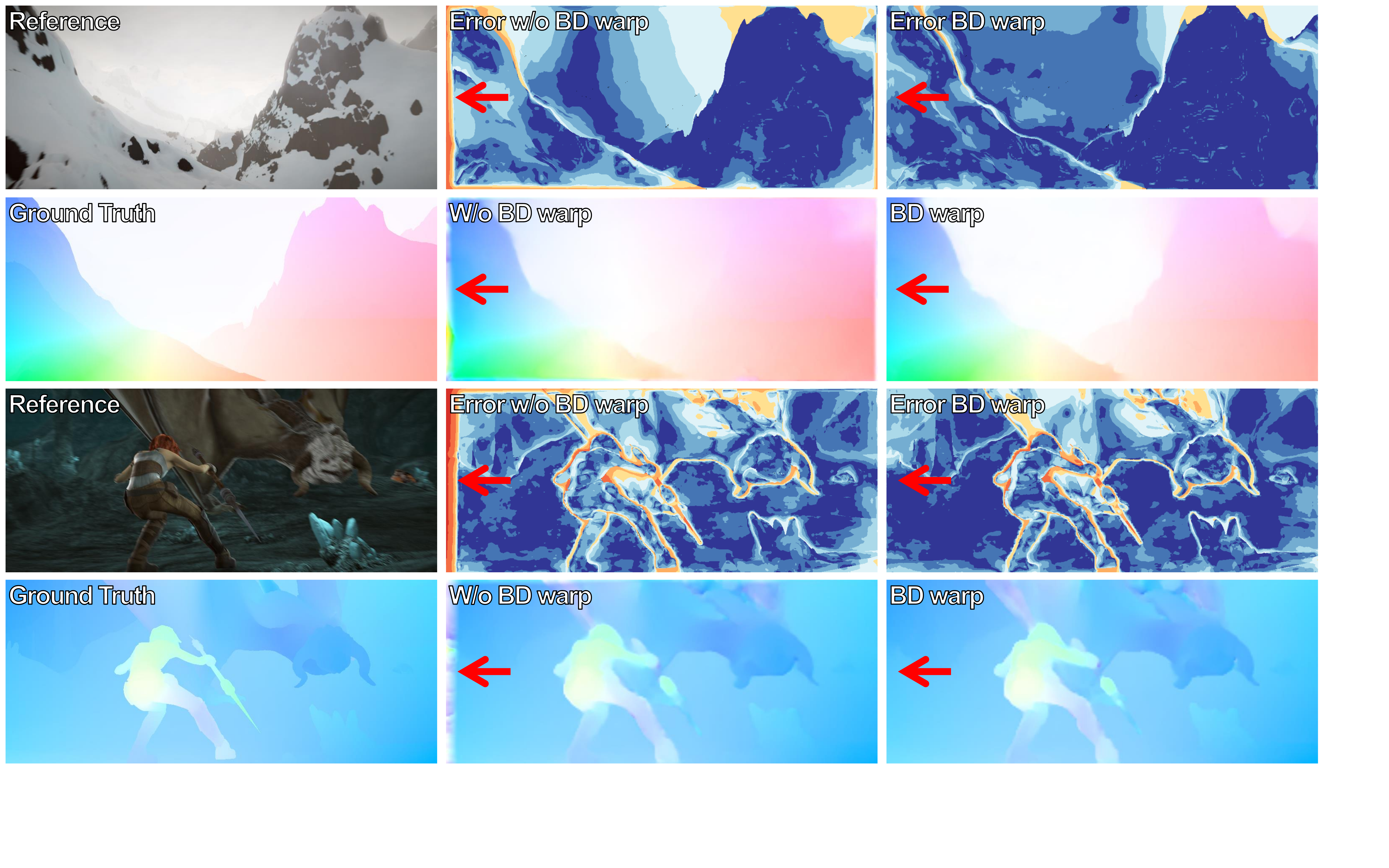}\label{fig:abl_result_if_bdwarp_Sintel_final}
  }
  \vspace{-3mm}
  \caption{Comparison of with v.s. without the boundary dilated warping on MPI-Sintel Clean (a) and Final (b) data sets. }\label{fig:abl_result_if_bdwarp}\vspace{-3mm}
\end{figure}

\vspace{-2mm}
We further show comparison results with our boundary dilated warping enabled or not (occlusion inpainting method is disabled at the same time) on multiple datasets such as Flying Chairs (Fig.~\ref{fig:abl_result_if_bdwarp_flying_chairs}), KITTI 2012 (Fig.~\ref{fig:abl_result_if_bdwarp_kitti2012}), Sintel Clean (Fig.~\ref{fig:abl_result_if_bdwarp_Sintel_clean}) and Sintel Final (Fig.~\ref{fig:abl_result_if_bdwarp_Sintel_final}) as well. As seen from the results, optical flow estimation in areas close to the image boundaries is well learned if our boundary dilated warping is involved.

\subsection{Multi-scale Analysis}\vspace{-2mm}
As detailed in Table~\ref{tab:ablation_study_multi_scale}, we conduct experiments of the occlusion inpainting on each single scale of our network, to reveal the effectiveness of the multi-scale design. 
As illustrated, occlusion inpainting achieves more gains on multi-scale than on single scale processing.

\begin{table}[!ht]
\centering
\resizebox*{\textwidth}{!}{
\begin{tabular}{
>{\centering\arraybackslash}p{3.4cm}
>{\centering\arraybackslash}p{1.9cm}
>{\centering\arraybackslash}p{1.9cm}
>{\centering\arraybackslash}p{1.9cm}
>{\centering\arraybackslash}p{1.9cm}
>{\centering\arraybackslash}p{1.9cm}}
\toprule
Refinement  & Chairs& KITTI 2012 & KITTI 2015 & Sintel Clean & Sintel Final \\
scale  & test&  train & train &  train &  train \\
\midrule
    w/o. OccInp  &   2.99    &    1.80   &   4.65   &   2.95   &    4.25   \\
$1/64$ scale    &   2.99    &    1.79   &   4.58   &   2.96   &    4.25   \\
$1/32$ scale    &   3.00    &    1.79   &   4.64   &   2.95   &    4.26   \\
$1/16$ scale    &   2.97    &    1.80   &   4.65   &   2.96   &    4.24   \\
$1/8$ scale     &   2.88    &    1.80   &   4.65   &   2.93   &    4.23   \\
$1/4$ scale     &   2.80    &    1.80   &   4.65   &   2.90   &    4.18   \\
\midrule
multi-scale (ours)  &   \textbf{2.62}    &    \textbf{1.78}   &   \textbf{4.57}   &   \textbf{2.82}   &    \textbf{4.13}   \\
\bottomrule
\end{tabular}}
\vspace{1mm}
\caption{Occlusion inpainting (OccInp) on each single scale. We report EPE values of our results on Flying Chairs, KITTI and MPI-Sintel datasets. }\label{tab:ablation_study_multi_scale}
\end{table}

\end{appendices}

\end{document}